\documentclass[lettersize,journal]{IEEEtran}
\usepackage{amsmath,amsfonts}
\usepackage{amssymb}
\usepackage{algorithmic}
\usepackage{algorithm}
\usepackage{array}
\usepackage[caption=false,font=normalsize,labelfont=sf,textfont=sf]{subfig}
\usepackage{textcomp}
\usepackage{stfloats}
\usepackage{url}
\usepackage{verbatim}
\usepackage{graphicx}
\usepackage{cite}
\usepackage{multirow}
\usepackage[colorlinks,bookmarksopen,bookmarksnumbered,citecolor=blue, linkcolor=blue, urlcolor=blue]{hyperref}
\usepackage[switch]{lineno}
\begin{document}

\title{Enhancing Scene Classification in Cloudy Image Scenarios: A Multi-modality Collaborative Transfer Method with Information Regulation Mechanism}

\author{Yuze Wang, Rong Xiao, Haifeng Li, Mariana Belgiu, Chao Tao$^*$


\thanks{The work presented in this paper was supported by the National Natural Science Foundation of China (No. 42171376); The Distinguished Young Scholars under Grant 2022JJ10072; The Open Fund of Xiangjiang Laboratory under Grant 22XJ03007.}

\thanks{Yuze Wang, Rong Xiao, Haifeng Li, Chao Tao are with the School of Geosciences and Info-Physics, Central South University, Changsha 410083, China}

\thanks{Mariana Belgiu is with the Faculty of Geo-Information Science and Earth Observation (ITC), University of Twente, Enschede,7500 AE, The Netherlands}}

\markboth{Journal of \LaTeX\ Class Files,~Vol.~14, No.~8, August~2021}%
{Shell \MakeLowercase{\textit{et al.}}: A Sample Article Using IEEEtran.cls for IEEE Journals}


\maketitle

\begin{abstract}
In remote sensing scene classification, leveraging the transfer methods with well-trained optical models is an efficient way to overcome label scarcity. However, cloud contamination leads to optical information loss and significant impacts on feature distribution, challenging the reliability and stability of transferred target models. Common solutions include cloud removal for optical data or directly using Synthetic aperture radar (SAR) data in the target domain. However, cloud removal requires substantial auxiliary data for support and pre-training, while directly using SAR disregards the unobstructed portions of optical data. This study presents a scene classification transfer method that synergistically combines multi-modality data, which aims to transfer the source domain model trained on cloud-free optical data to the target domain that includes both cloudy optical and SAR data at low cost. Specifically, the framework incorporates two parts: (1) the collaborative transfer strategy, based on knowledge distillation, enables the efficient prior knowledge transfer across heterogeneous data; (2) the information regulation mechanism (IRM) is proposed to address the modality imbalance issue during transfer. It employs auxiliary models to measure the contribution discrepancy of each modality, and automatically balances the information utilization of modalities during the target model learning process at the sample level. The transfer experiments were conducted on simulated and real cloud datasets, demonstrating the superior performance of the proposed method compared to other solutions in cloud-covered scenarios. We also verified the importance and limitations of IRM, and further discussed and visualized the modality imbalance problem during the model transfer. Codes are available at https://github.com/wangyuze-csu/ESCCS
\end{abstract}



\section{Introduction}
\label{sec1}
\IEEEPARstart{R}{emote} sensing image scene classification tasks play a pivotal role in supporting environmental monitoring \cite{liu2023multi}, resource management \cite{ghazouani2019multi}, urban planning \cite{liu2017classifying}, and military reconnaissance \cite{su2023reconstruction}, attracting extensive attention across the field. \textcolor{red}{With the continuous refinement of application requirements, dynamic scene monitoring has evolved from annual assessments to more frequent intervals, such as quarterly or even monthly monitoring \cite{cheng2020remote}. Given the high susceptibility of optical remote sensing data acquisition to weather conditions, cloud contamination often affects images captured during specific phases. According to the International Satellite Cloud Climatology Project-Flux Data (ISCCP-FD), the global monthly cloud cover is approximately 66.5\% \cite{zhang2004calculation,mendoza2021thermodynamics}. Therefore, it is challenging to acquire cloud-free optical images across various periods, thereby hindering effective dynamic scene monitoring.} This phenomenon is especially noticeable in tropical, subtropical, and coastal regions, where prolonged rainy seasons substantially hinder the ability to acquire cloud-free images over long periods \cite{yang2019object}.

As deep learning models and transfer learning methods continue to develop, they collectively offer novel solutions for the efficient and low-cost completion of scene classification tasks \cite{niu2020decade}. Deep learning models enable the automated extraction of sophisticated features, which can efficiently enhance their scene recognition capabilities \cite{cheng2020remote}. Transfer learning methods, most commonly via fine-tuning \cite{niu2020decade,wang2022empirical}, make well-trained deep learning models rapidly adaptable to new scene recognition tasks with few labeled data by fully leveraging prior knowledge. However, cloud coverage degrades the information within optical images by obscuring spatial structure, texture, and context, and also significantly affects the scene's feature distribution. This degradation impacts the performance of fine-tuning the models well-trained in cloud-free optical images, such as NWPU-RESISC45 \cite{cheng2017remote} and AID \cite{xia2017aid}, to target scenes with cloud-affected optical images. \textcolor{red}{Moreover, given the strong capability of SAR images to penetrate clouds and their high sensitivity to the geometric structure of land covers, SAR images become a reliable source for scene classification when optical information is missing. Despite these advantages, the complexity of labeling\cite{ding2017learning} and the significant spatio-temporal diversity \cite{huang2020classification} of SAR images pose considerable challenges in the pretraining of models and their deployment in local applications.} Consequently, many pre-trained-transfer methods are hard to utilize in cloud-prone regions effectively. For the above problem of scene classification with cloud contamination, the following two main solutions are currently available:

The first solution is to transfer the model trained on the cloud-free optical images to the cloud-cover optical images that have been processed with the cloud removal method. For example, Lorenzi et al \cite{lorenzi2011inpainting} proposed reconstructing the cloud-covered regions by propagating the texture structure of the cloud-free background, and further optimizing the results by enriching the feature space and search range of the background information. Such methodologies are susceptible to the issue of edge effects. With the development of deep learning models, Xu et al. \cite{xu2021missing} have introduced an edge generation network to improve the scene realism of the reconstructed areas, and leverage the model's strong perception of depth and spatial relationships to improve its robustness across varied types of cloud coverage. Considering the serious information deficit caused by thick cloud coverage, Zhang et al. \cite{zhang2018missing} developed a unified spatial-temporal-spectral model for thick cloud removal, which can fully utilize the auxiliary information from multi-source and multi-temporal data to accurately reconstruct the cloud-corrupted regions. Given the difficulty of obtaining auxiliary information in some areas due to climatic conditions,  Tao et al. \cite{tao2022thick} proposed a texture complexity-guided self-paced learning (SPL) framework to construct the thick cloud-covered regions from a single image, which can be adapted to diverse de-clouded scenes and achieved more detailed texture restoration in the data-deficient scenarios. 

While the cloud removal methods have shown good performance in areas with limited thick cloud coverage or slightly affected by thin clouds, they still face challenges in cloud-prone regions. Firstly, the absence of cloud-free images over neighborhood periods and the extensive cloud cover in some regions deprives the model of the necessary information needed for accurately restoring scenes beneath the cloud \cite{gawlikowski2022explaining}. Additionally, these methods are limited by their generalizability, which are usually designed and trained for specific scenes. When the thickness, texture, and shape of clouds are substantially changed within the scenes, the models may show unstable performance, leading to distortion and edge effects \cite{shen2014effective}. To adapt to the target cloud-covered images, they often require a substantial amount of additional data to keep recognition and restoration abilities, which increases both the time and manual costs in cloud-prone regions.

The second solution is to transfer the model trained on cloud-free optical images to Synthetic Aperture Radar (SAR) images \cite{liu2018can}. Nevertheless, the significant modality gap between SAR and optical data remains a major challenge in transferring well-trained optical models \cite{huang2020classification}. Researchers usually alleviate the domain gap through the methods at the sample and feature levels. At the sample level, they typically explore the correlations between SAR and optical data to enable the conversion from SAR to optical images, thereby fully leveraging the scene recognition capabilities of the source optical model. For example, Wang et al. \cite{wang2022sar} used Parallel-Generative Adversarial Network (Parallel-GAN) to reconstruct the optical images from SAR images. Song et al. \cite{song2022two} employed Cycle-GAN \cite{senapati2023image} to transform labeled optical images from the source domain into images with SAR style, and re-trained the source model to enhance its capability to recognize features from SAR images, which makes it easier to transfer the source model to the target scenes. At the feature level, researchers often narrow the modality gap by mapping different data to a cross-modality shared feature space. For example, Rostami et al. \cite{rostami2019deep} trained two deep encoders to map the optical and SAR images into a latent embedding space, and force the model to minimize the distribution distance between the two modalities. \textcolor{red}{Zhu et al.\cite{zhu2021deep} and Peng et al.\cite{peng2022domain} extracted specific layer activation from the encoders of both the source and target models, aligning them with metric learning strategies to constrain the target model's learning process.} These approaches can convert the feature extraction capabilities obtained in the optical source domain model to the SAR target domain, which enables effective utilization of the prior knowledge from the source domain model. 

In real-world scenarios, the thickness and coverage of clouds across different regions within optical images exhibit considerable variability, which makes the regions covered by thin clouds or partially obscured by thick clouds still retain abundant information. \textcolor{red}{Moreover, the characteristics of SAR data alone may not suffice to distinguish certain scenes with a high degree of similarity, such as built-up \cite{attarchi2020extracting} and vegetation cover \cite{dos2021vegetation}. Optical information can provide additional spectral, textural, and local spatial structural details to SAR data, thereby enhancing the separability of different scenes within the feature space\cite{amarsaikhan2010fusing,bai2021comprehensively}. Consequently, leveraging the synergistic and complementary effects of optical and SAR data can enhance the detection and discrimination of features for scene classification tasks in cloud-prone regions.}

\begin{figure}[t]
	\begin{center}
        \includegraphics[width=1\linewidth]{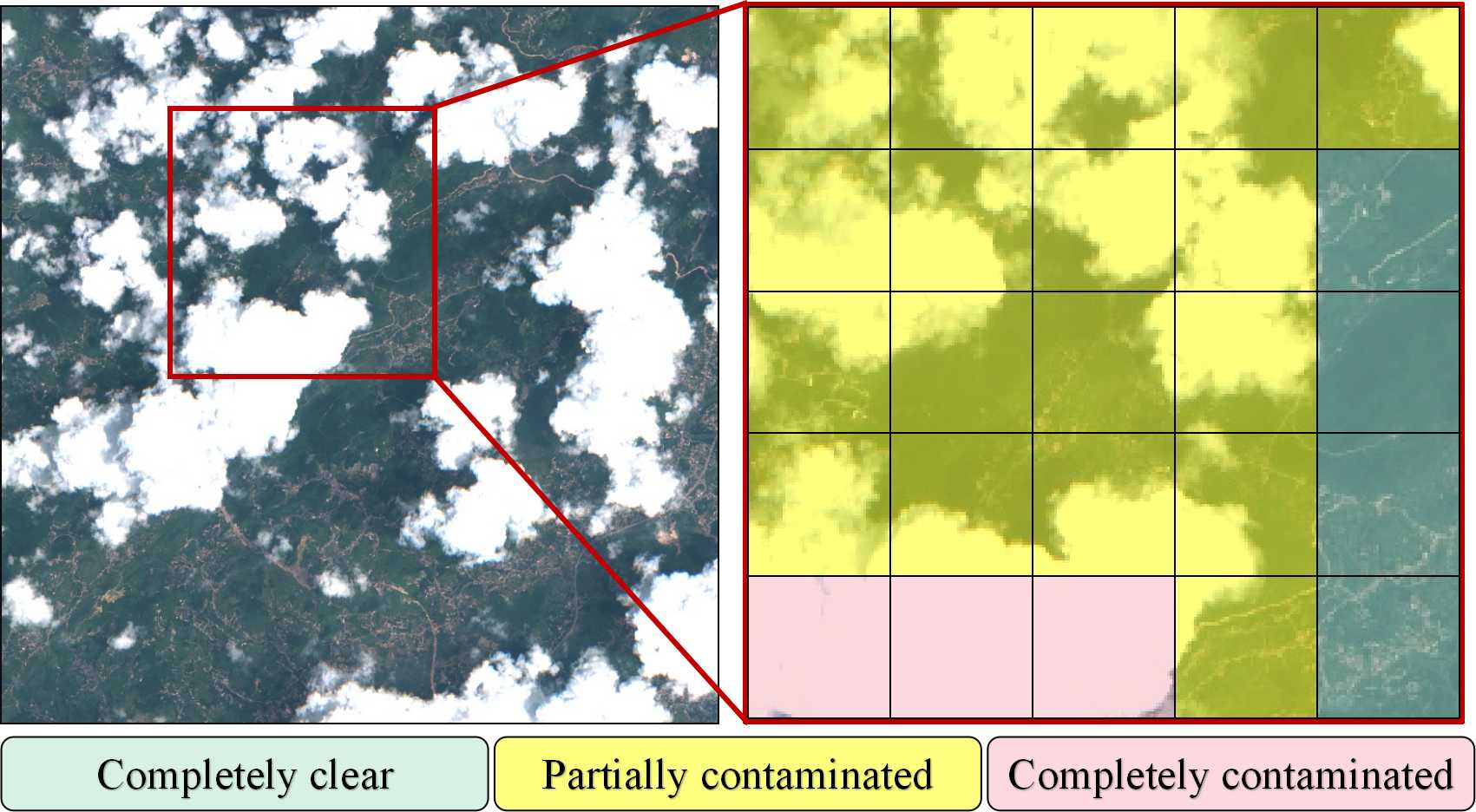}
		\caption{\textcolor{red}{Valid optical information remains in the yellow part (partially contaminated).}}
		\label{F1}
	\end{center}
\end{figure}

\textcolor{red}{Building on this rationale, we attempt to effectively utilize optical information in cloud-prone regions where cloud contamination is frequent, particularly when cloud cover is between completely clear and completely contaminated, as illustrated in the partially contaminated areas highlighted in yellow in Fig.\ref{F1}.} Such optical images are often difficult to restore using cloud removal methods due to the lack of auxiliary data and high costs \cite{tao2022thick}. Entirely abandoning optical data and relying solely on SAR data would forgo a significant amount of valuable information that could potentially enhance the model's capabilities \cite{mahyoub2019fusing}. This study proposes a collaborative transfer strategy for scene classification in cloud-prone regions, enabling the transfer of prior knowledge from the cloud-free optical source domain to the target domain containing both cloud-covered optical data and SAR data. \textcolor{red}{Specifically, based on knowledge distillation \cite{wang2021knowledge}, this strategy not only achieves transfer between heterogeneous data using pseudo labels \cite{wang2021cross} but also utilizes unlabeled samples to improve the model's generalization ability. Additionally, we construct auxiliary domains for each modality within the target domain to further integrate and optimize the prior knowledge from the source model, thereby alleviating the significant domain gap between the source and target domains \cite{tao2023general}.}

However, during the observation of the transfer process, we noted that the direct transfer could lead to a 'modality imbalance' issue, which is caused by the distinct domain gap between the source domain data and the multi-modality data within the target domain. Specifically, during the transfer process, the target model tends to fit the modality that is easier to fit, which means the model will overly rely on the modality with lower transfer difficulty (superior modality). The target model tends to disregard the rich information contained in the modality that is hard to fit, which means the model will suppress the modality with higher transfer difficulty (inferior modality). For instance, in scenarios where the target scene is covered by a thin cloud or a small part thick cloud, the domain gap between cloud-cover optical target images and source data is smaller than the domain gap between SAR target images and source data. Notably, when target domain optical images are heavily or completely contaminated by clouds, the target model may tend to fit the SAR images. Since the main purpose of our study is to effectively harness residual information within optical images that are partially cloud-affected, extreme cases with severe cloud contamination offer minimal information to enhance scene recognition for our method. Hence, we have not further discussed such scenarios, and believe that directly employing SAR images might be a more suitable way when optical images are heavily cloud-contaminated.

Focusing on the modality imbalance problem, we have further designed a data-driven Information Regulation Mechanism (IRM) within our collaborative transfer framework. Specifically, we initially construct auxiliary models for optical and SAR modalities respectively. This not only concretizes the superior and inferior states of different modalities during the transfer process, but also integrates and optimizes the prior knowledge from the source model. Subsequently, we dynamically balance the contributions of each modality at the sample level within the target model's decision-making process, which enables the model to adapt to various cloud-affected scenarios. \textcolor{red}{It enhances the model's attention to valuable information on inferior modality and prevents overfitting on the superior modality, which can fully exploit the synergistic and complementary potential of multi-modality data to achieve a '$1+1 > 1$' effect.}
To validate the proposed method, this study conducts experiments with optical-SAR remote sensing scene classification datasets under both simulated and real cloud cover conditions. 

To summarize, the main contributions are as follows:

\begin{enumerate}
	\setlength\itemsep{0em}\setlength\parskip{0em}\setlength\topsep{0em}\setlength\partopsep{0em}\setlength\parsep{0em} 
        \vspace{5pt}
	\item{The collaborative transfer method is proposed to utilize a pair of cloud-covered optical and SAR images to enhance the performance of the scene classification target model in cloud-prone regions, which achieves the best performance compared to other solutions under both simulated and real cloud-cover conditions.} 
        \vspace{5pt}
	\item{Focused on the problem of modality imbalance. We designed a data-driven IRM to dynamically adjust and couple the information of each modality. The modality imbalance problem is further visualized, and the importance and limitations of IRM are discussed. }
        \vspace{5pt}
        \end{enumerate}

The remainder of this paper is organized as follows: Section.\ref{sec1} introduces the framework and key components of the proposed method. In Section.\ref{sec2}, we introduce the process for constructing the collaborative transfer and information regulation mechanism. Section.\ref{sec3} describes the experimental settings and presents the main results compared to other methods. In Section.\ref{sec4}, we discuss the modality imbalance problem and the limitations of the proposed method. Finally, we draw some conclusions and introduce future work in Section.\ref{sec5}. 
\begin{figure*}[!t]
	\begin{center}
        \includegraphics[width=1\linewidth]{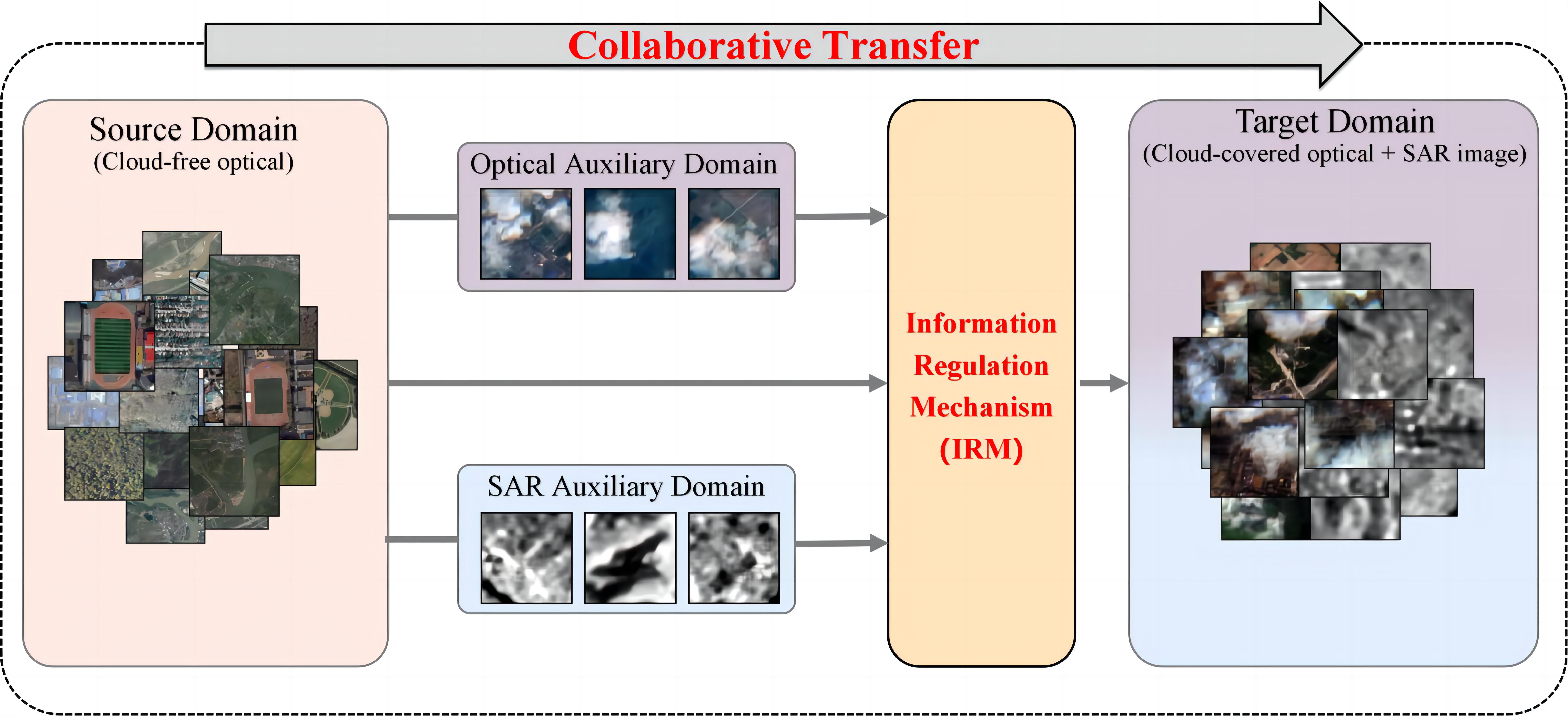}
		\caption{The general framework of the proposed method. 
It primarily consists of two components: the Collaborative Transfer Strategy and the Information Regulation Mechanism. The former is designed to facilitate efficient transfer between heterogeneous data, while the latter tackles the issue of modality imbalance.}
		\label{F2}
	\end{center}
\end{figure*}

\section{Methodology}
\label{sec2}

\subsection{General framework}
\label{subsec2.1}
\indent

Addressing the challenge of scene classification tasks under cloud-affected conditions, this paper integrates optical images partially disturbed by clouds and SAR images as the target domain, and designs a transfer method to efficiently utilize the multi-modality information. Specifically, as shown in Fig.\ref{F2}, the whole framework is combined in two parts: (1) \textit{Collaborative transfer strategy}: based on the knowledge distillation \cite{gou2021knowledge}, we build the multi-step transfer framework between heterogeneous data with the assistance of auxiliary models, thereby flexibly integrating the abundant information from both optical and SAR modalities. The auxiliary models also serve to concretize the comparative advantages and disadvantages of different modalities throughout the transfer process. (2) \textit{Information regulation mechanism (IRM)}: by calculating the contribution value of each modality based on the logit outputs from the auxiliary models, we enable the model to learn the ability to handle various modality imbalance states. After balancing the superior and inferior modalities, the model can more effectively leverage multi-modality data's synergistic and complementary effects.

Before detailing the proposed method, we introduce the following symbols and terms for clarity: Given the cloud-free optical images $X_S$ and corresponding labels $Y_S$, we refer to the source domain as $S = \{X_S, Y_S\}$, and the target domain with cloud-covered optical and SAR image pairs referred to as $T = \left[X_{T_{Opt}}, X_{T_{SAR}}, Y_T\right]$. Considering the challenge of acquiring labels in cloud-prone regions, our target domain consists of a limited number of labeled samples alongside an abundance of unlabeled samples. The model is pre-trained with source data to obtain the $f_{pre}$, and the $f_{pre}$ is fine-tuned to accommodate the distinct class scheme between the source and target domains to obtain the source model $f_S$.

\begin{figure*}[!t]
	\begin{center}
        \includegraphics[width=0.9\linewidth]{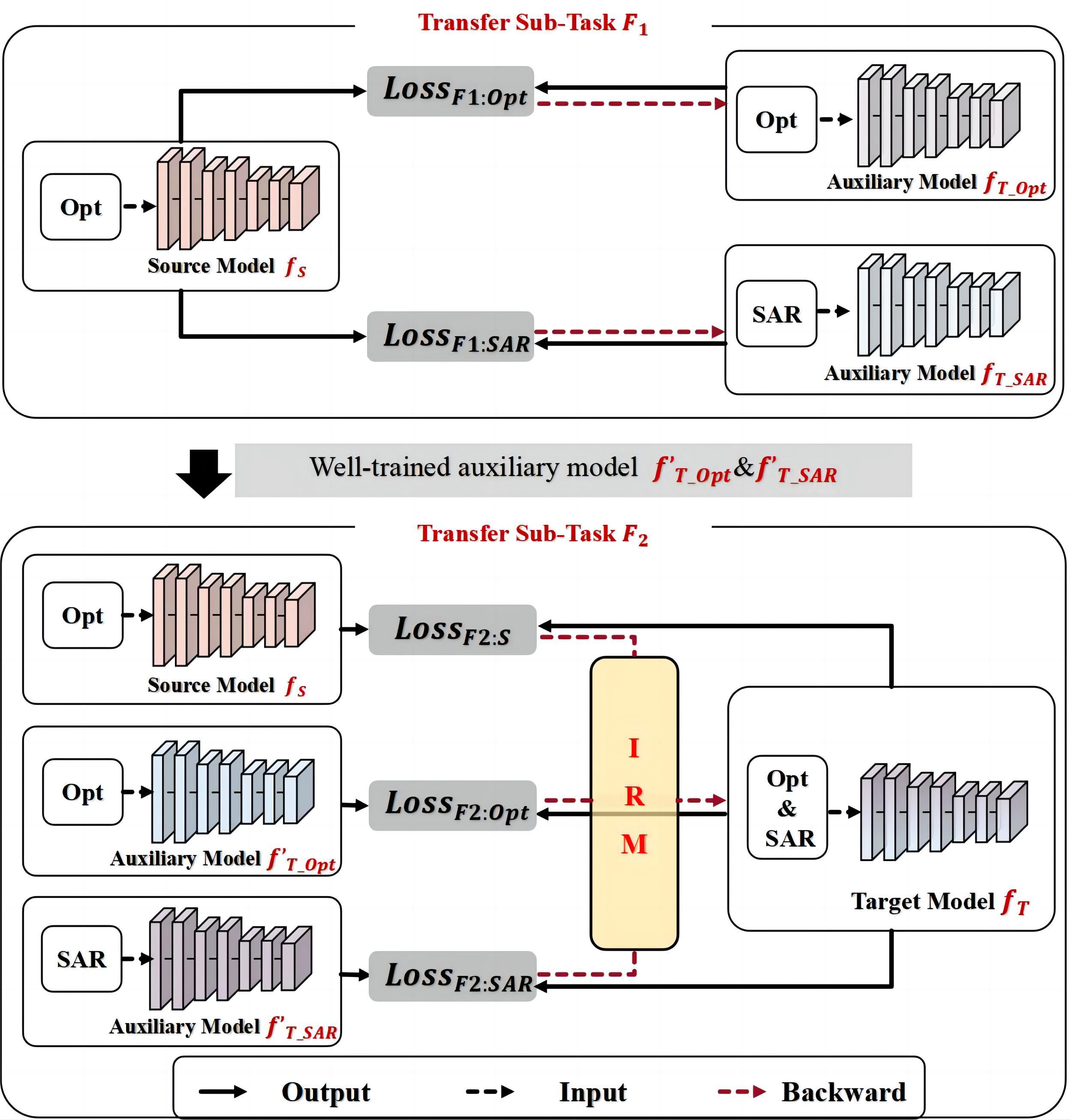}
		\caption{The collaborative transfer steps of the proposed method.}
		\label{F3}
	\end{center}
\end{figure*}

\subsection{Collaborative transfer using optical-SAR remote sensing image pairs}\label{subsec2.2}
\indent
During the collaborative transfer, we take the pseudo-labels as the bridge to transfer prior knowledge between source and target domains. Firstly, it not only enables the transfer process between heterogeneous data, but also further integrates and optimizes prior knowledge from the source domain \cite{wang2021cross}. Secondly, it can fully utilize the massive unlabeled images to enrich the feature representation space during the model learning process, which can enhance the model’s generalization ability to recognize diverse scenarios \cite{wang2022sar}. Lastly, and most critically, it is helpful to concretize the modality's contribution. The logit representation of pseudo-labels conveys significant cues, serving as an informational foundation for assessing modality contributions. As shown in Fig.\ref{F3}, the collaborative transfer is consisting of two sub-tasks:
\vspace{5pt}

\noindent {1) The first transfer sub-task ($F_1$): From source model to auxiliary models}

\vspace{5pt}
Given the optical images from the target domain as input for the source and optical auxiliary models, the SAR images from the target domain serve as inputs to the SAR auxiliary model. \textcolor{red}{The source model $f_S$ is used to generate the pseudo-labels $Y = f_S(X_{T_{Opt}})$, while the predictions $Q_{T_{Opt}} = f_{T\_Opt}(X_{T_{Opt}})$ and $Q_{T_{SAR}} = f_{T\_SAR}(X_{T_{SAR}})$ are produced by the optical/SAR auxiliary models $f_{T\_Opt}/f_{T\_SAR}$, separately}. Since the source model has been well-trained on the cloud-free datasets, the pseudo-labels generated by $f_S$ guide the training of the two auxiliary models as supervision signals. \textcolor{red}{Specifically, the optical/SAR regulation model $f_{T\_Opt}/f_{T\_SAR}$ will be guided by using the cross-entropy loss $Loss_{F1:Opt}/Loss_{F1:SAR}$ calculated by the pseudo-labels $Y_S$ and their prediction results $Q_{T_{Opt}}/Q_{T_{SAR}}$:}

\begin{equation}\label{equ1}
Loss_{F1:Opt} = -\sum_{i=1}^{N} Y_{S,i} \log(Q_{T_{Opt},i})
\end{equation}

\begin{equation}\label{equ2}
Loss_{F1:SAR} = -\sum_{i=1}^{N} Y_{S,i} \log(Q_{T_{SAR},i})
\end{equation}

Two auxiliary models can be well-trained in the first transfer sub-task, which can provide integrated supervision signals and concretize the modality's contribution for the second training step.

\vspace{5pt}
\noindent 2) The second transfer sub-task ($F_2$): From source/auxiliary models to target model:
\vspace{5pt}

We take the multi-modality images as input for the auxiliary models and the target model. \textcolor{red}{The two trained auxiliary models \(f'_{T\_Opt}\) and \(f'_{T\_SAR}\) generate the pseudo-labels \(Q'_{T_{Opt}} = f'_{T\_Opt}(X_{T_{Opt}})\) and \(Q'_{T_{SAR}} = f'_{T\_SAR}(X_{T_{SAR}})\), separately.} The predictions \(Q_T = f_T(X_T)\) are generated by the target model \(f_T\) through optical and SAR images simultaneously. Next, we combine the pseudo-labels from two single-modality as the supervision signals to guide the target model training. The source model is also employed as one of the teacher models, acting as a regularization term to prevent the amplification of erroneous prior knowledge during the previous step. Combined with the above supervision signals, the cross-entropy loss to train the target model \(Loss_T\) is calculated:
\begin{equation}\label{equ3}
	\begin{aligned}
		Loss_T =& Loss_{F2:Opt} + Loss_{F2:SAR} + Loss_{F2:S} \\
		=&-\sum_{i=1}^{N} Q'_{T_{Opt},i} \log(Q_{T,i}) \\ 
		&- \sum_{i=1}^{N} Q'_{T_{SAR},i} \log(Q_{T,i}) \\ 
		&- \sum_{i=1}^{N}  Y_{S,i} \log(Q_{T,i})  \\
    \end{aligned}
\end{equation}

\subsection{Information regulation mechanism for modality imbalance problem}\label{subsec2.3}

Based on the Principle of Least Effort (PLE) \cite{geirhos2020shortcut}, the model prefers to learn the data that can be more easily fitted during the learning process. When this phenomenon extends to the learning process of multi-modality data, it results in the model tends to over-fitting the modalities that are easier to comprehend, while underfitting the modalities that are more challenging to understand \cite{wang2020makes}, which is also known as modality imbalance problem. Multi-modality learning (MML) focuses on exploiting the different information across modalities to jointly address their inherent limitations. This problem leads to a significant deficiency in the model's learning of inferior modality, thereby preventing the full utilization of synergistic and complementary information across different modalities \cite{peng2022balanced}. In this study, we further extend the problem of modality imbalance to the transfer process from the single-modality source domain to the multi-modality target domain. The factors leading to modality imbalance have been transformed into differing domain gaps between the source domain modality and different target domain modalities:
\begin{equation}
g(S,T_1) \neq g(S,T_2) \neq \cdots \neq g(S,T_n)
\end{equation}
where $g(\cdot)$ represents the domain gaps between the source domain S and sub-target domains with different modalities \( \{T_1, T_2, \ldots, T_n\} \). In the task of transferring prior knowledge from a cloud-free optical source domain to a target domain composed of cloud-affected optical and SAR data, the magnitude of domain gaps is jointly influenced by the differences in imaging mechanism, learning difficulty in various scenarios \cite{chauhan2016comparative}, and degree of cloud contamination.

\begin{figure*}[!t]
	\begin{center}
        \includegraphics[width=1.9\columnwidth]{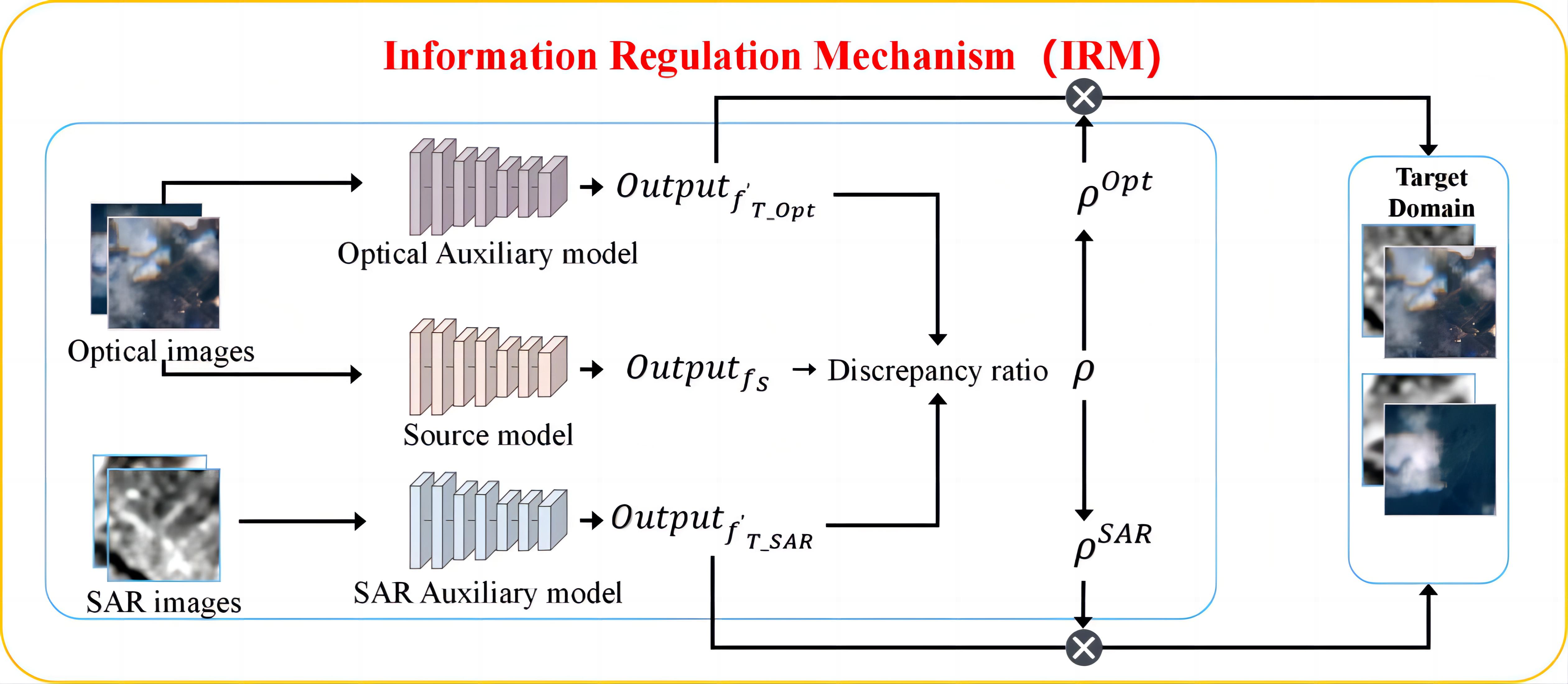}
		\caption{Architecture of the Integrated Representation Model (IRM).}
		\label{F4}
	\end{center}
\end{figure*}

To address this problem, inspired by \cite{peng2022balanced}, we propose a data-driven IRM that automatically adjusts the relative prominence of different modalities. This mechanism leverages the auxiliary models from the previous section to calculate the contributions of each modality as the base information source. During the adjustment process, information from inferior modalities is given increased attention, while temporarily reducing the focus on information from superior modalities, ensuring that all modalities can be adequately learned. 

Furthermore, the comparative advantages and disadvantages between the two modalities vary across different cloud coverage levels and scenarios. For instance, as the cloud cover decreases, the domain gap between source and optical target domains is gradually reduced, and the model will exhibit a greater bias towards learning from optical data. In contrast, due to the imaging mechanism, SAR and optical data show similar features in scenarios without significant height changes, such as bare land or grassland. This similarity makes SAR data easier to interpret, and the model tends to be less biased towards optical data. Therefore, we extend the IRM to the sample level, allowing it to dynamically modulate the status of modality imbalance based on the relative relationship between each pair of optical and SAR images. Besides, we have also incorporated the source model to jointly guide the learning process of the target model. It not only prevents the accumulation of errors during the knowledge distillation, but also serves as a regulation item to avoid an excessive bias of the model toward inferior modality. The detailed process is shown in Fig.\ref{F4}.

Firstly, we introduce the concept of the information contribution value $S$ to quantify the significance of the information embedded within optical and SAR images. This metric is calculated based on the logits output from auxiliary models $f'_{T\_Opt}$ and $f'_{T\_SAR}$, serving as a measure of the importance of the information provided by each modality:

\begin{align}
S^{Opt} &= \sum_{k=1}^{M} 1_{k=Q_{T\_Opt}} \cdot softmax(Output_{f'_{T\_Opt}})_k, \\
S^{SAR} &= \sum_{k=1}^{M} 1_{k=Q_{T\_SAR}} \cdot softmax(Output_{f'_{T\_SAR}})_k,
\end{align}
where $M$ is the number of categories in the dataset, and $Q_{T\_Opt}$, $Q_{T\_SAR}$ are the predicted class indices by the auxiliary models for each sample.

Next, we dynamically monitor the contribution discrepancy between optical and SAR modalities through the calculation of a discrepancy ratio $\rho$:

\begin{align}
\rho^{Opt} &= \frac{\sum_{i \in B} S_i^{SAR}}{\sum_{i \in B} S_i^{Opt}}, &
\rho^{SAR} &= \frac{\sum_{i \in B} S_i^{Opt}}{\sum_{i \in B} S_i^{SAR}}.
\end{align}
where $B$ represents the batch size during model input. In the modulation process, our goal is to guide the model's attention more towards the inferior modality and reduce focus on the superior one to prevent information dependency. We use $\rho$ as weight factors to influence the target model's loss $Loss_T$:

\begin{equation}
	\begin{aligned}
		Loss_T = &\text{IRM}\left(Loss_{F2:Opt} + Loss_{F2:SAR} + Loss_{F2:S}\right) \\
		=& \rho^{Opt} \cdot Loss_{F2:Opt} + \rho^{SAR} \cdot Loss_{F2:SAR} + Loss_{F2:S} \\
		=& -\rho^{Opt} \cdot \sum_{i=1}^{N} Q'_{T\_{Opt,i}} \log(Q_{T,i}) \\
		&- \rho^{SAR} \cdot \sum_{i=1}^{N} Q'_{T\_{SAR,i}} \log(Q_{T,i})  \\
		& - \sum_{i=1}^{N}  Y_{S,i} \log(Q_{T,i})
	\end{aligned}
\end{equation}

\noindent where the weight of the supervised signal from the source model is fixed to prevent over-adjustment or overfitting. By adjusting the IRM, the target model can fully leverage the rich information from multi-modality data, maximizing the synergistic and complementary potential between modalities.

\section{Experiment}
\label{sec3}

\subsection{Experiment settings}\label{subsec3.1}

\subsubsection{Datasets}\label{subsec3.1.1}

\begin{figure*}[!t]
	\begin{center}
        \includegraphics[width=1.8\columnwidth]{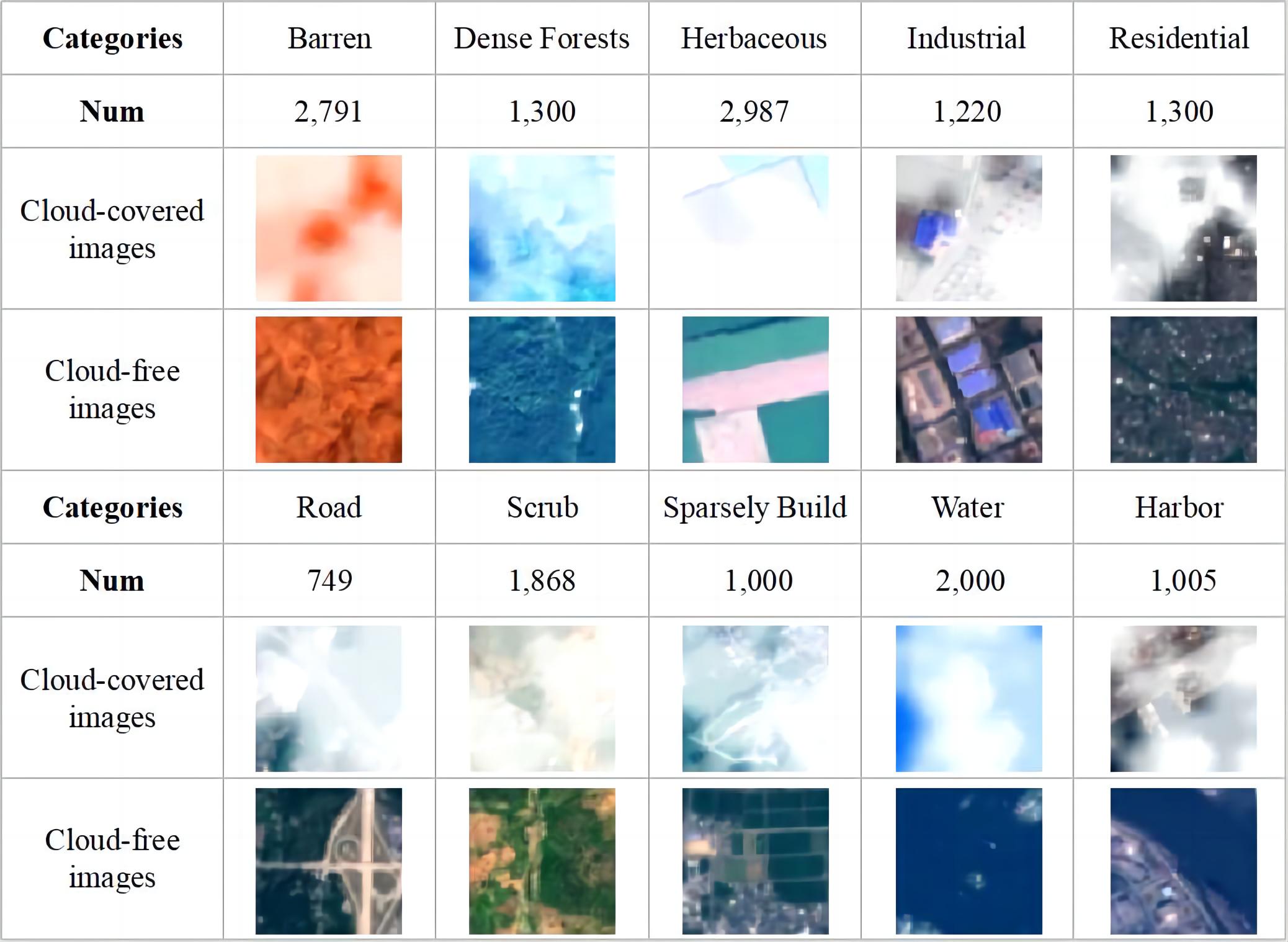}
		\caption{Numbers and example of each category in SEN12MS Cloud dataset.}
		\label{F5}
	\end{center}
\end{figure*}

\indent
The proposed framework was validated by two cloud-covered RS scene classification datasets, including a simulated cloud-covered and a real cloud-covered RS images classification dataset.  The source models are pre-trained by the NWPU-RESISC45(NR) and AID datasets.
\vspace{5pt}

\textbf{The simulated cloud-covered images dataset (SEN12MS Cloud):}We reconstruct a scene classification dataset, named ‘SEN12MS Cloud’, based on an existing public RS images dataset ‘SEN12MS’ \cite{schmitt2019sen12ms}, which contains 16,219 pairs of corresponding SAR and optical images with the size of 64 ×64 pixels. The SAR images are collected from the Sentinel-1 satellite covering 3 channels (VV, VH, and VV/VH) with resolutions of 10m/pixels. The optical images are collected from the Sentinel-2 satellite covering 3 spectral bands (R, G, and B) with resolutions of 10m/pixel. \textcolor{red}{Based on observations from real cloud-cover images, we found that optical images are not uniformly contaminated by clouds in actual scenes. A common situation is that some of the samples remain completely cloud-free after cropping. Firstly, to simulate real-world application conditions more accurately, we applied real cloud masks \cite{tao2022thick} to 50\% of the samples at the image level, resulting in a total of 8,110 samples, with each sample having an approximate cloud masking rate of 50\%. Secondly, to better reflect real-world conditions, we incorporated thin and thick cloud masks during the process, maintaining an approximate 1:1 ratio between them. These pairs of images are divided into 10 categories. Some samples of this dataset are shown in Fig.\ref{F5}.}

\begin{figure*}[!t]
	\begin{center}
        \includegraphics[width=1.8\columnwidth]{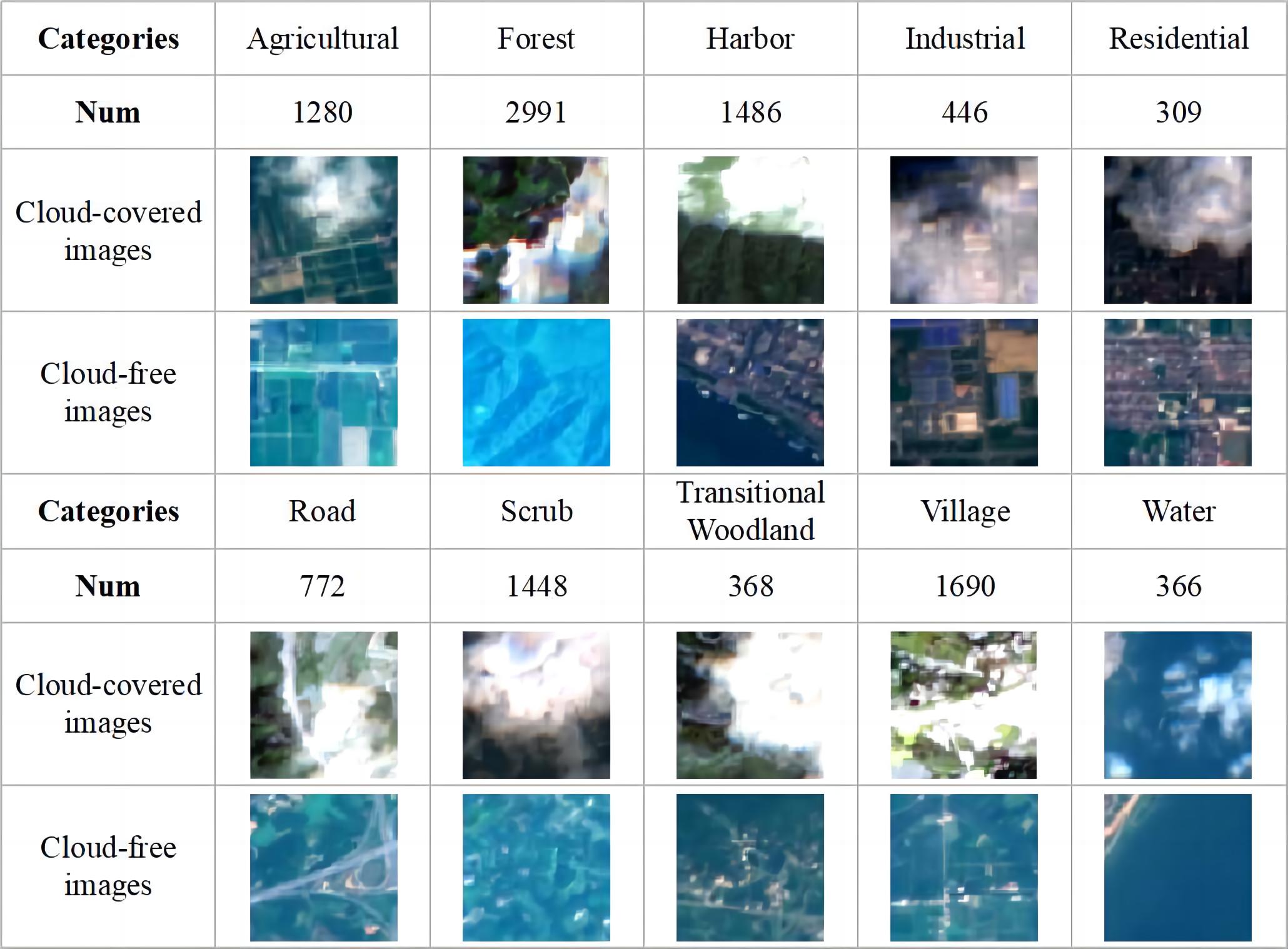}
		\caption{Numbers and example of each category in Hunan Cloud dataset.}
		\label{F6}
	\end{center}
\end{figure*}

\vspace{5pt}
\textbf{The real cloud-covered images dataset (Hunan Cloud):} \textcolor{red}{We have developed a dataset for real cloud-covered scene classification, which was collected Sentinel-2A and Sentinel-2B Level-2A Bottom of Atmosphere reflectance images (S2-L2A) for optical data, and Sentinel-1 Level-1 Ground Range Detected (GRD) for SAR data, covering the northern region of Hunan Province in July 2018.} This region is located in the subtropical climate belt and is frequently influenced by rainy/cloudy weather, especially in the summer \cite{zhang2018mapping}. The image data of this dataset was collected from Sentinel-1(VV, VH, and VV/VH) and Sentinel-2 satellite (R, G, and B) sensors with resolutions of 10m/pixel in July 2018. We ultimately selected 11,156 pairs of corresponding SAR and cloud-covered optical images with a size of 64×64 pixels. The cloud cover accounts for approximately 70\% of the total content in all optical images. Additionally, we divide all samples into 10 categories, and some samples of this dataset are shown in Fig.\ref{F6}.

\vspace{5pt}
\textbf{The cloud-free optical datasets for the source domain:}The source models are pre-trained by the NR and AID datasets, which are widely used for scene classification. The NR dataset comprises 31,500 aerial images across 45 categories, with RGB bands. The images are of size 256x256 pixels, with a spatial resolution of 30–0.2 m/pixel. The AID dataset consists of 10,000 aerial images categorized into 30 classes, with RGB bands. The images are of size 600x600 pixels and have a spatial resolution of 8-0.5 m/pixel.

\subsubsection{Training details}\label{subsec3.1.2}

\indent
In our experiments, we employed an 80/20\% random split for training and testing the model. Within the training set, we utilized 20 labeled samples per category to fine-tune the source model, adjusting it to the target class scheme. The remaining unlabeled samples were employed for knowledge distillation. Additionally, during the experimental process, we used Overall Accuracy (OA), Average Accuracy (AA), and Kappa as the primary metrics for evaluation.

We set the learning rate as $1 \times 10^{-3}$ and the learning rate decay factor as $5\times 10^{4}$. We set the training epoch as 200 and the batch size as 64 for the two datasets. The experiments were conducted with PyTorch, the platform on the Ubuntu 18.04 operation system, and three NVIDIA GTX3080 GPUs with 11 GB memory. For all experiment settings, we used the widely applied 'ResNet-50' \cite{he2016deep}  as the source and took the 'GoogLeNet' \cite{szegedy2016rethinking} as the target network \cite{wang2021cross}. 

\subsubsection{Compared methods}\label{subsec3.1.4}

\textcolor{red}{We have employed three types of methods for comparative analysis with our proposed approach. These methods exhibit different adaptability for the modalities from the target domain, which can be classified into:(1) transfer from optical to optical modality (Opt to Opt); (2) transfer from optical to SAR modality (Opt to SAR); and (3) transfer from optical to both optical and SAR modalities (Opt to Opt\&SAR). All methods utilized only 20 labeled samples or sample pairs.}

\vspace{5pt}
\textbf{Opt to Opt}: \textcolor{red}{We utilized three methods, including Finetune, KD-S, and SPL. Finetune involved freezing the complete encoder and retraining the classifier \cite{wang2022empirical}. KD-S applied knowledge distillation using soft pseudo-labels to convey prior knowledge \cite{wang2021knowledge}. SPL incorporated a Self-paced Learning strategy for cloud removal before final classification \cite{tao2022thick}. The pre-trained cloud remove model is modified by 20 cloud/cloud-free sample pairs to adopt the target scenarios.}

\vspace{5pt}
\textbf{Opt to SAR}: \textcolor{red}{We employed four methods including Finetune, CycIT, TTL, and DSAN. Finetune involved freezing parts of the encoder and retraining the classifier \cite{wang2022empirical}. CycIT used CycGAN \cite{senapati2023image} to translate SAR images into an optical style before final classification \cite{song2022two}. The pre-trained translation model is modified by 20 optical-SAR sample pairs to adopt the target scenarios. TTL utilized the Transitive Transfer Learning strategy to construct an intermediate domain to bridge the domain gap between different modalities \cite{tao2023general}. DSAN is a method that employs unsupervised domain adaptation strategies at the feature level to mitigate the domain gap between different modalities \cite{peng2022domain,zhu2021deep}.}

\vspace{5pt}
\textbf{Opt to Opt\&SAR}: \textcolor{red}{We utilized KDHN, TLF, and SL three methods. KDHN employs knowledge distillation to facilitate knowledge transfer across heterogeneous networks, filtering samples with high prediction uncertainty during the transfer process \cite{wang2021cross}. TLF uses a late fusion strategy during transfer, extracting features from different modalities using separate encoders and combining them for classification \cite{li2022deep}. SL employs a supervised learning strategy to directly learn multi-modality information.}

\subsection{Main results}\label{sec3.2}

\subsubsection{Experiment I: The Simulated Cloud-covered Images Experiments}\label{sec3.2.1}

\indent
Table \ref{Tabb1} and Table \ref{Tabb2} present the transfer results on the simulated cloud-covered images dataset for our method along with the comparison methods mentioned above. The source domain was set to NR and AID, and the target domain was set to SEN12MS Cloud. 

When utilizing only optical data from the target domain, the Finetune method demonstrates inferior performance due to the information loss induced by cloud contamination. Specifically, when transferring from the NR and AID datasets, the accuracy for the Finetune method is 47.04\% and 55.06\%. For the SPL method, which employs the cloud removal model to restore the cloud-covered images, shows no significant improvement over the Finetune method and may have a negative impact. For instance, when AID and NR serve as the source domain, the SPL method results in less than a 1.00\% increase and a 9.57\% decrease in OA compared to the Finetune method. This is attributed to the fact that cloud removal methods often require extensive cloud and cloud-free pairs from the target scene to ensure their reliability. In the absence of such pairs due to climatic conditions, the cloud removal method might fail to entirely eliminate clouds or produce spurious artifacts, thereby leading to misleading classifications by the model. The KD-S performed well, with the first- and second-best OA achieved when using the NR and AID datasets as source domains in compared methods, respectively, due to its ability to learn diverse features from unlabeled data, enhancing robustness for cloud-covered scenarios.

\begin{table}[!t]
	\renewcommand{\arraystretch}{1.6}
    \setlength{\tabcolsep}{10pt}
	\centering
 	\caption{\textcolor{red}{Performance of proposed and compared methods for transferring from NR to SEN12MS Cloud dataset.}}
	\begin{tabular}{cccc}
\hline
Methods             & OA(\%)         & AA(\%)         & Kappa          \\ \hline
Finetune (Opt only) \cite{wang2022empirical} & 47.04          & 59.77          & 0.379          \\
SPL \cite{tao2022thick}                & 42.54          & 47.17          & 0.356          \\
KD-S \cite{wang2021knowledge}              & 59.38          & 62.38          & 0.531          \\
Finetune (SAR only) \cite{wang2022empirical} & 22.64          & 22.06          & 0.140          \\
CycIT \cite{senapati2023image,song2022two}              & 30.59          & 28.86          & 0.221          \\
TTL \cite{tao2023general}                & 26.84          & 10.12          & 0.113          \\
\textcolor{red}{DSAN} \cite{peng2022domain,zhu2021deep}               & 48.52          & 20.70          & 0.395          \\
SL                  & 42.54          & 43.53          & 0.344          \\
KDHN \cite{wang2021cross}                & 68.60          & 69.63          & 0.640          \\
\textcolor{red}{TLF} \cite{li2022deep}                & 69.31          & 65.89          & 0.654          \\
\textbf{Ours}       & \textbf{78.50} & \textbf{76.64} & \textbf{0.754} \\ \hline
\end{tabular}
	\label{Tabb1}
\end{table}

\begin{table}[!t]
	\renewcommand{\arraystretch}{1.6}
    \setlength{\tabcolsep}{11pt}
	\centering
 	\caption{\textcolor{red}{Performance of proposed and Compared methods for transferring from AID to SEN12MS Cloud dataset.}}
        \begin{tabular}{cccc}
        \hline
        Methods             & OA(\%)         & AA(\%)         & Kappa          \\ \hline
        Finetune (Opt only) \cite{wang2022empirical} & 55.06          & 53.87          & 0.483          \\
        SPL \cite{tao2022thick}                   & 55.34          & 62.11          & 0.482          \\
        KD-S \cite{wang2021knowledge}               & 55.27          & 61.95          & 0.481          \\
        Finetune (SAR only) \cite{wang2022empirical} & 19.62          & 20.12          & 0.103          \\
        CycIT \cite{senapati2023image,song2022two}              & 17.52          & 16.59          & 0.124          \\
        TTL \cite{tao2023general}                & 23.84          & 10.90          & 0.094          \\
        \textcolor{red}{DSAN} \cite{peng2022domain,zhu2021deep}                & 41.80          & 40.06          & 0.308          \\
        SL                  & 42.54          & 43.53          & 0.344          \\
        KDHN \cite{wang2021cross}                & 63.02          & 65.39          & 0.574          \\
        \textcolor{red}{TLF} \cite{li2022deep}                  & 62.92          & 59.60          & 0.583          \\
        \textbf{Ours}       & \textbf{75.17} & \textbf{75.83} & \textbf{0.716} \\ \hline
	\end{tabular}
	\label{Tabb2}
\end{table}

When utilizing only SAR data from the target domain, all compared methods demonstrate poor performance, with OA below 50\%. Specifically, the performance of Finetune and TTL methods is significantly hindered by the substantial domain gap between SAR and optical modalities. \textcolor{red}{Although DSAN alleviates this issue by aligning features at the feature level, the substantial differences in their imaging mechanisms still impede the transfer of prior information. The CycIT method, which involves the translation of SAR data from the target domain to optical styles, faces challenges due to the limited availability of cloud-free optical and SAR image pairs for training. As a result, the model struggles to establish accurate mapping relationships between optical and SAR images, ultimately leading to distorted images and significant noise in the translated data.}

\begin{table}[t]
	\renewcommand{\arraystretch}{1.6}
    \setlength{\tabcolsep}{11pt}
        \caption{\textcolor{red}{Performance of proposed and Compared methods for transferring from NR to Hunan Cloud dataset.}}
	\centering
	\begin{tabular}{ccccc}
        \hline
        Methods             & OA(\%)         & AA(\%)         & Kappa          \\ \hline
        Finetune (Opt only) \cite{wang2022empirical} & 36.67          & 33.25          & 0.285          \\
        SPL \cite{tao2022thick}                   & 22.62          & 26.02          & 0.139          \\
        KD-S \cite{wang2021knowledge}               & 42.73          & 38.90          & 0.343          \\
        Finetune (SAR only) \cite{wang2022empirical} & 46.32          & 34.61          & 0.372          \\
        CycIT \cite{senapati2023image,song2022two}              & 21.44          & 21.07          & 0.123          \\
        TTL \cite{tao2023general}                   & 13.69          & 12.06          & 0.120          \\
        \textcolor{red}{DSAN} \cite{peng2022domain,zhu2021deep}                 & 60.46          & 49.34          & 0.525          \\
        SL                  & 40.62          & 34.35          & 0.313          \\
        KDHN \cite{wang2021cross}                 & 44.57          & 38.38          & 0.361          \\
        \textcolor{red}{TLF} \cite{li2022deep}                   & 52.69          & 49.68          & 0.431          \\
        \textbf{Ours}       & \textbf{68.00} & \textbf{55.08} & \textbf{0.623} \\ \hline
	\end{tabular}

	\label{Tabb3}
\end{table}

\begin{table}[!t]
	\renewcommand{\arraystretch}{1.6}
    \setlength{\tabcolsep}{9pt}
        \caption{\textcolor{red}{Performance of proposed and compared methods for transferring from AID to Hunan Cloud dataset.}}
	\centering
	\begin{tabular}{ccccc}
        \hline
        Methods             & OA(\%)         & AA(\%)         & Kappa          \\ \hline
        Finetune (Opt only) \cite{wang2022empirical} & 36.27          & 33.55          & 0.284          \\
        SPL \cite{tao2022thick}                & 23.65          & 27.80          & 0.161          \\
        KD-S \cite{wang2021knowledge}                 & 40.35          & 35.26          & 0.317          \\
        Finetune (SAR only) \cite{wang2022empirical} & 44.70          & 35.91          & 0.362          \\
        CycIT \cite{senapati2023image,song2022two}              & 19.50          & 21.13          & 0.119          \\
        TTL \cite{tao2023general}                & 12.21          & 14.22          & 0.134          \\
        \textcolor{red}{DSAN} \cite{peng2022domain,zhu2021deep}                & 60.68          & 46.41          & 0.526          \\
        SL                  & 40.62          & 34.35          & 0.313          \\
        KDHN \cite{wang2021cross}               & 40.57          & 35.64          & 0.321          \\
        \textcolor{red}{TLF} \cite{li2022deep}                 & 42.73          & 40.60          & 0.353          \\
        \textbf{Ours}       & \textbf{65.08} & \textbf{49.57} & \textbf{0.583} \\ \hline
	\end{tabular}

	\label{Tabb4}
\end{table}

When simultaneously utilizing optical and SAR data from the target domain, the SL method encounters challenges in effectively capturing the relationship between optical and SAR modalities with limited labeled samples. The Finetune method is not suitable for heterogeneous data. \textcolor{red}{Consequently, we have adopted the TLF approach to fusing multi-modality features, enabling it to extract independent features from different modalities and integrate them for joint decision-making. However, since TLF still lacks interaction between the multi-modality information during the learning process, it only achieves an OA of 69.31\% and 62.92\% when used as a source domain in NR or AID, respectively. Furthermore, although KDHN overcomes the limitations of the model's architecture and makes full use of abundant unlabeled data, enabling the model to extract information from different modalities and learn their interrelations, the concurrent input of multi-modality data also leads to a reliance on superior modality and neglect of the inferior ones. This has resulted in KDHN's performance being only comparable to TFL's.} Our approach further addresses this issue during the transfer process by incorporating a collaborative transfer framework with the Information Regulation Mechanism (IRM), which can effectively leverage synergistic and complementary information from both optical and SAR modalities. This is evidenced by a substantial 13.26\% and 19.28\% improvement over the best-performed method in terms of OA, while also outperforming methods solely reliant on optical cloud-covered data or SAR data.

\begin{figure*}[!t]
	\begin{center}
        \includegraphics[width=1.9\columnwidth]{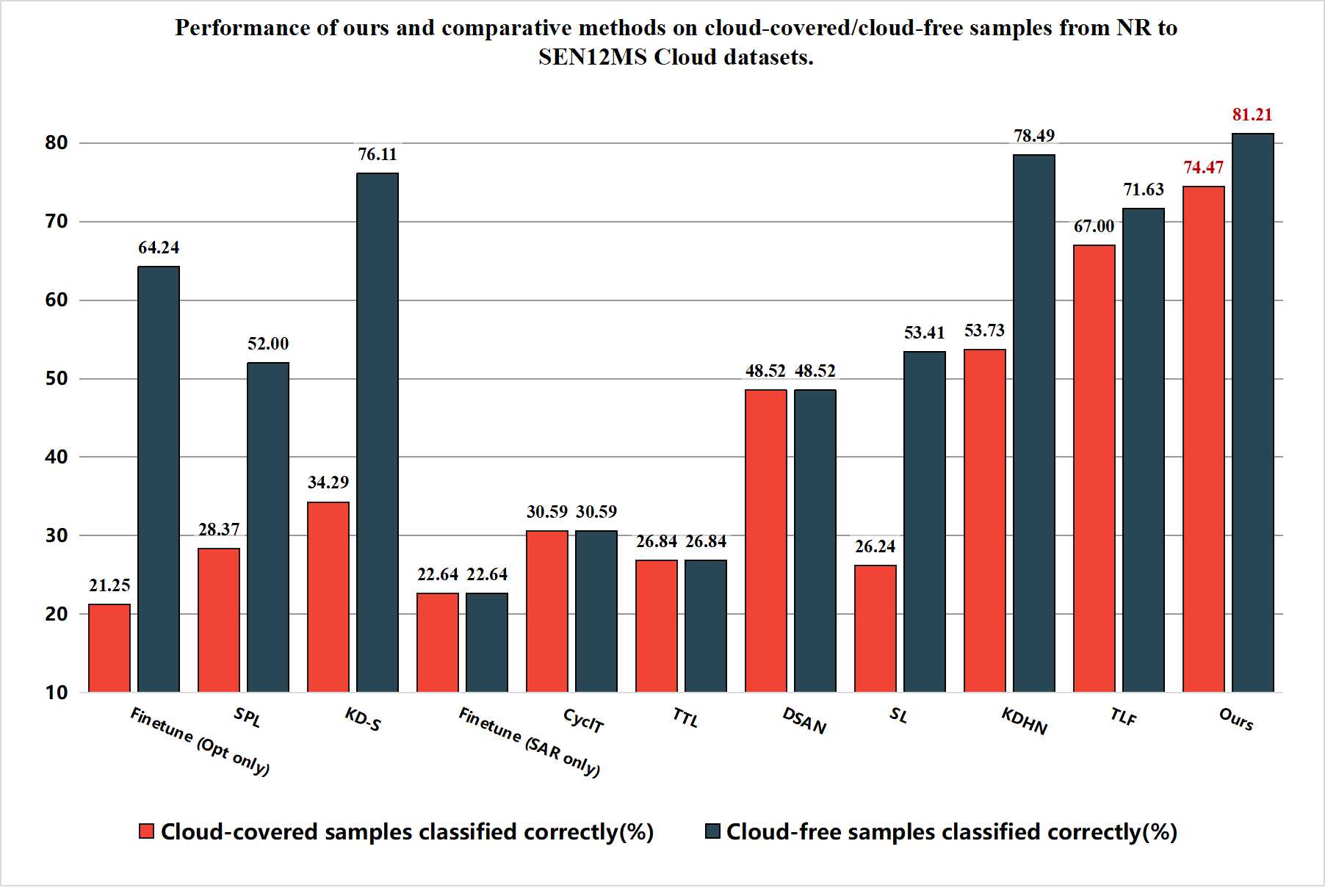}
		\caption{\textcolor{red}{Performance of proposed and comparative methods on cloud-covered/cloud-free samples from NR to SEN12MS Cloud Datasets. Among these, Finetune(Opt only), SPL, and KD-S are methods that are applicable only to the optical modality. Finetune(SAR only), CycIT, TTL, and DSAN are methods that are applicable only to the SAR modality, hence their performance is consistent across cloud-covered and cloud-free samples. SL, KDHN, and TLF are methods capable of utilizing both optical and SAR modalities.}}
		\label{F67}
	\end{center}
\end{figure*}

\subsubsection{Experiment II: The Real Cloud-covered Images Experiments}
\label{sec3.2.2}
\indent
To further demonstrate the effectiveness of our method in real-world scenarios, we conduct the same experiment on the real cloud-covered images dataset. The source domain was set to NR and AID. The target domain was set to the Hunan Cloud dataset, which contains much higher cloud-cover ratios and complex cloud-cover conditions compared to the simulated dataset. 

As shown in Table \ref{Tabb3} and Tabel \ref{Tabb4}, the SL and all Finetune methods demonstrate poor performance due to the lack of labeled training samples, with OA ranging from 34.78\% to 46.32\%. The KD-S also faces challenges under conditions of significant loss of optical information. Meanwhile, the SPL and CycIT methods suffer severe negative effects on transfer performance due to a lack of cloud-free/cloud-covered optical and optical/SAR image pairs in more complex target domains, yielding OA below 25.00\%. The KDHN method, utilizing AID and NR datasets as source domains, achieves only 44.57\% and 40.57\% in OA, respectively. The complexity of cloud coverage and scene features amplifies the negative impact of modality imbalance, leading the model to over-fit to the superior modality and leaving it unable to supplement information from the inferior modality. 

\textcolor{red}{A noteworthy observation is that nearly all methods, including ours, exhibit a general decline in performance relative to their performance on simulated datasets. Conversely, DSAN shows superior results to the simulated dataset, reaching an OA of 60.46\% and 60.68\% across various source domains. Based on the data observation, the primary cause of this phenomenon appears to be the substantial presence of hilly terrain in the real dataset from the Hunan region, which facilitates DSAN's ability to effectively align optical and SAR data at the feature level. Nevertheless, compared to DSAN, our method still outperforms DSAN, achieving improvements in OA of 12.47\% and 7.25\% using NR and AID as source domains, respectively. This further emphasizes the benefits of our method in addressing modality imbalance, potentially enhancing the robustness of scene classification models in real cloud-prone regions.}

\begin{figure*}[!t]
	\begin{center}
        \includegraphics[width=1.8\columnwidth]{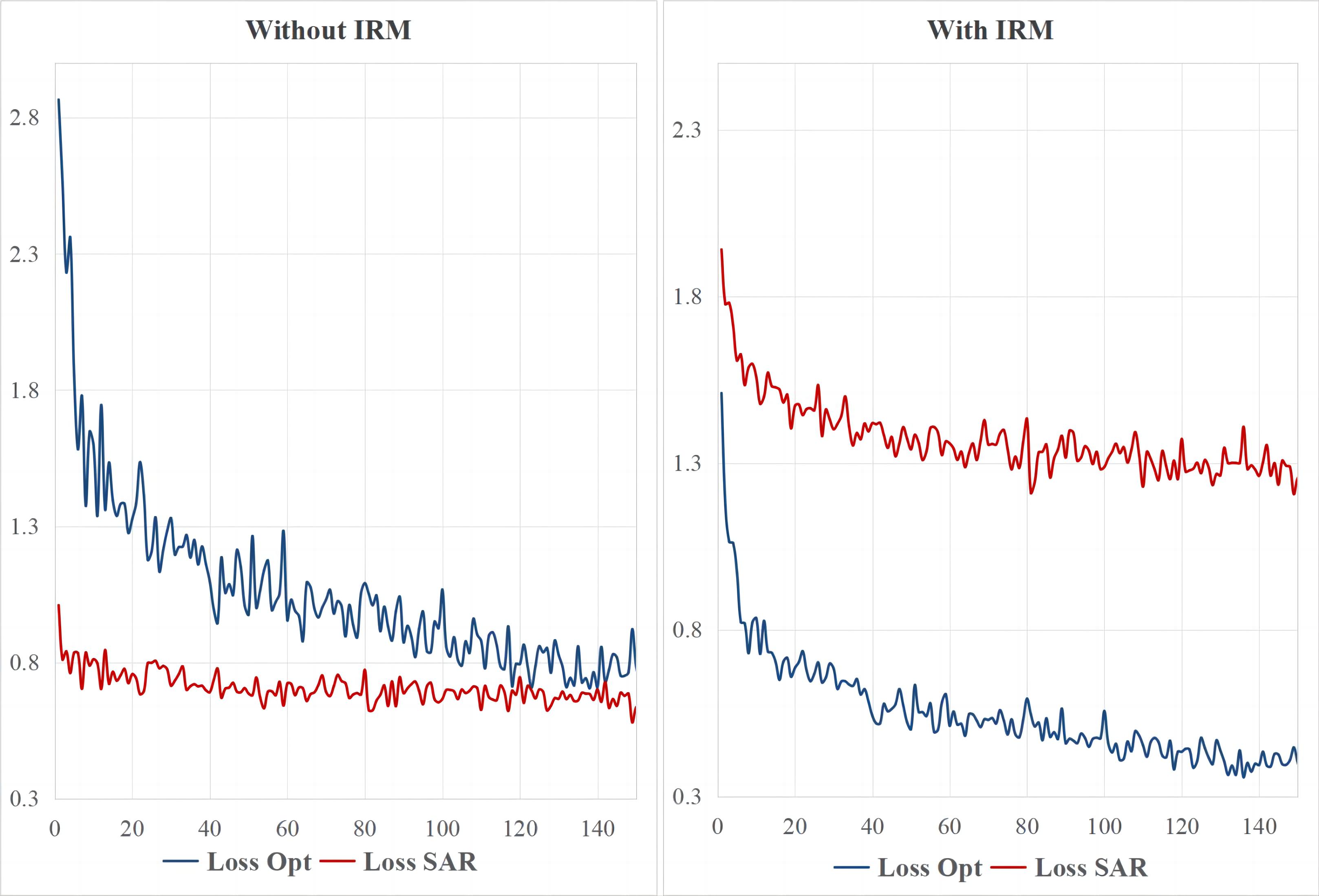}
		\caption{Visualizing model’s loss to optical/SAR modality before and after incorporating IRM, with the red and blue lines representing the loss values on optical and SAR data, respectively. }
		\label{F8}
	\end{center}
\end{figure*}

\subsection{Ablation experiments}
\label{sec3.3}
\indent
Given that this study primarily focuses on scenarios where clouds are present but are not fully obscure, the datasets are designed to contain both cloud-covered and cloud-free samples. Although our approach demonstrates significant enhancements in overall performance, its specific improvements on cloud-covered and cloud-free samples remain unclear. Additionally, as mentioned before in Section \ref{subsec2.3}, the issue of modality imbalance also exists within cloud-free data pairs, and the presence of clouds further complicates this issue. Therefore, to assess the effectiveness of our proposed method on both cloud-covered and cloud-free samples, we statistically analyzed the accuracies of the proposed and compared methods on cloud-covered and cloud-free samples separately. This analysis also serves to further investigate the impact of cloud contamination on the modality imbalance issue during the transfer process. All the compared methods are still categorized based on their applicability to the data modalities from the target domain. The source domain is chosen as the NR dataset, and the target domain is the SEN12MS Cloud dataset. 

As shown in Fig.\ref{F8}, for the methods applicable only to optical modality, it is observed that they achieve good results on cloud-free imagery, particularly the KD-S method, but their performance on cloud-covered images decreased. Despite being unaffected by clouds, methods applied for the SAR modality exhibit poor performance due to the significant domain gap with the source domain's optical data. With the simultaneous use of optical and SAR modalities, the KDHN and TLF methods have yielded substantial improvements over the method that solely relies on a single modality in cloud-covered samples, which nearly doubles the accuracy. This improvement is attributed to the heterogeneous data transfer capability of the knowledge distillation \cite{wang2021cross} and feature fusion \cite{li2022deep}, which provides a substantial basis enabling the model to exploit synergies and complementarities between multi-modality data \cite{li2022deep}.  

Although the overall performance of the multi-modality method exceeds the single-modality method, the accuracy for cloud-covered samples is still hard to meet practical application requirements. This happens because cloud contamination disrupts the spectral distribution and spatial structure of optical images \cite{gawlikowski2022explaining}, which may complicate the modality imbalance during the transfer process. Our proposed method can automatically help the model regulate inter-relationships between modalities during transfer, and improve the model's utilization of diverse modalities, which significantly increases the accuracy for cloud-covered images by 11.15\%. Additionally, it not only significantly improves cloudy images but also enhances the accuracy of cloud-free images by 3.46\% compared with the best-performance method.

We believe that the reason our proposed framework yields greater benefits for cloud-covered samples than for cloud-free samples is the complexity of modality imbalance. When optical images are not disrupted by clouds, the imbalance in learning multi-modality information is primarily determined by the similarity between the representation of imaging mechanisms for different scene contexts \cite{chauhan2016comparative}. In such cases, the model can autonomously learn to handle these relatively simple situations through extensive observation of data \cite{huang2021makes}. When cloud contamination emerges as a factor to impact the modality imbalance, the complexity increases due to the varying extent, thickness, and shape of cloud covers, which makes the model hard to handle such cases with its robustness. The regulation of the proposed method enables the model to gain the ability to automatically balance the learning process in complex cases, and the efficiency of multi-modality information utilization in cloud-free images is also improved.

\section{Discussion}
\label{sec4}
In this section, we conducted a series of experiments to comprehensively analyze the modality imbalance problem, while further exploring the advantages and limitations of information regulation mechanism (IRM). Specifically, our discussions are centered on two main aspects: visualizing and examining the modality imbalance during transfer, and evaluating the efficacy and limitations of IRM across varying cloud content scenarios.

\begin{figure*}[!t]
	\begin{center}
        \includegraphics[width=1.9\columnwidth]{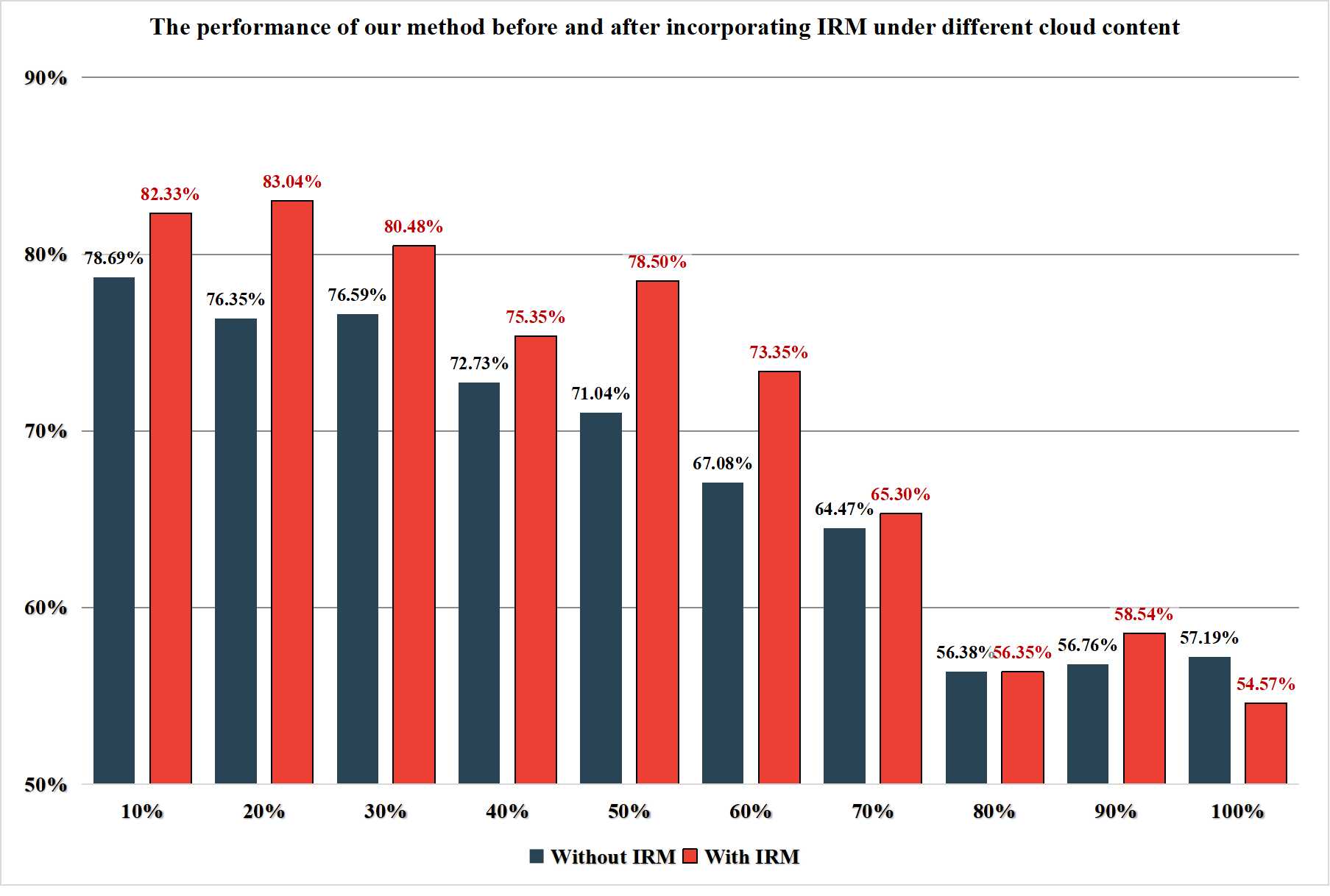}
		\caption{The performance of our method before and after incorporating IRM under different cloud content, with the inclusion of simulated cloud masks ranging from 10\% to 100\% of all samples.}
		\label{F9}
	\end{center}
\end{figure*}

\subsection{Analysis of the modal imbalance problem}
\label{4.1}

\indent
As indicated in Guo et al. \cite{gou2021knowledge} and Fan et al. \cite{fan2023pmr}, modality imbalance is a common issue across various domains. It is prevalent in several learning paradigms, including supervised learning \cite{wang2023imbalance}, unsupervised learning \cite{yang2019deep}, and self-supervised learning \cite{liu2021contrastive}. The primary cause of this situation is that models tend to learn from the data that can be more easily fitted during the learning process, which is typically represented by the variation in the loss value \cite{sun2021learning}. Therefore, to further analyze the imbalance problem and the effectiveness of the proposed IRM addressing it in the transfer process, we separately monitored the loss value for each modality and visualized the results before and after adding the IRM. Specifically, we tracked the loss of the respective modality over 150 epochs. The target domain is selected as the SEN12MS Cloud dataset, and the source domain is selected as the NR dataset.

As shown in Fig.\ref{F8}, before incorporating IRM, the model's loss on the optical modality exhibits a rapidly decreasing trend, indicating a notable fitting to optical information. The model’s loss on the SAR modality is almost unchanged, implying a limited utilization process of SAR information. The different trends of loss value changes across different modalities clearly represent the modality imbalance problem during the learning process. The superior modality not only suppresses the learning rates of other modalities, but also may interfere with their update direction \cite{fan2023pmr}, which makes it hard to jointly exploit the common priors of different modalities to overcome their inherent limitations.

Through IRM's dynamic adjustment of modality balance, it can be observed that the loss value exhibits a consistent relative pace of reduction, reflecting a synchronized trend throughout the optimization process. Notably, while the loss values across different modalities may not yet be entirely consistent in absolute terms, the model already optimizes them in a balanced way \cite{li2023boosting}, as absolute alignment might lead to missing specific information \cite{jiang2023understanding}. The model can further effectively utilize the information from the inferior modality to complement the superior modality, while potentially preventing the model from over-fitting the superior modality. Ultimately, this allows the model to better leverage the synergistic and complementary effects of multi-modality data, enhancing robustness in cloud-covered scenes.

\subsection{The limitation of IRM under different cloud content}
\label{sec4.2}
\indent
Our previous experiments were conducted in an ideal environment where only half of the optical images were obscured by clouds, leaving other parts completely cloud-free. However, real-world scenarios frequently show varied cloud-cover patterns within areas \cite{bony2015clouds}, with changing proportions of cloud-free and cloud-covered data. The motivation for the Information Regulation Mechanism (IRM) is to effectively utilize optical information in regions prone to cloud cover, enhancing models' scene classification abilities. However, our approach may have limitations when there is a substantial loss of optical information. To investigate the effectiveness and limitations of the IRM mechanism in practical application scenarios, we simulated various levels of cloud content from 10\% to 100\%, and statistically analyzed the model performance before and after implementing the IRM. For instance, to simulate a 10\% cloud cover scenario, we masked 1620 out of 16,219 images in our dataset, and the masks were derived from actual cloud images. The target domain is selected as SEN12MS Cloud datasets with different simulated cloud content, and the source domain is selected as NR dataset.

As shown in Fig.\ref{F9}, the method with IRM performs well with cloud content ranging from 10\% to 70\%, effectively addressing modality imbalance issues and enhancing our framework's performance by 1.29\% to 10.50\%. However, in extreme conditions where cloud content exceeds 70\%, the IRM demonstrates limited performance. The reasons are as follows: (1) Given that the proposed method is based on a knowledge distillation framework, it still needs a certain proportion of cloud-free samples to learn the accurate mapping relationship between optical and SAR modalities \cite{dou2020unpaired}. (2) Insufficient cloud-free samples lead to a substantial increase in the error rate of pseudo-labels during the model's knowledge distill transfer process. This significantly hinders the effective transfer of prior knowledge to scene recognition capabilities from the source domain model to the target model. (3) When most optical samples are clouded, the model lacks appropriate target images for learning the dynamic balance relationship between optical and SAR modalities. This can lead the model to overfit extreme adjustment states, consequently causing an ineffective automatic adjustment function.

In summary, the proposed IRM is only available in the cloud-prone regions that still retain optical information, which enhances the accuracy and frequency of monitoring in those application scenarios. For regions predominantly covered by clouds, our method is limited. Retraining the model with SAR images \cite{qian2021hybrid} or employing a special model transfer method \cite{rostami2019deep} might be more effective alternatives in such scenarios.

\section{Conclusion}
\label{sec5}
\indent
In this paper, we develop a scene classification transfer method for cloud-prone regions, which utilizes pairs of cloud-covered optical images and SAR images to mitigate the impact of cloud interference. The method effectively leverages the prior knowledge embedded in the source model, thereby enhancing the frequency and accuracy of monitoring in regions where cloud-free images are rarely available over extended periods, such as the summer season in subtropical monsoon climate areas. Based on the observation, the direct transfer of models between the source and target domains may be hindered by the 'modality imbalance' issue, which arises from the domain gap between multi-modality data in the target domain and optical data in the source domain. It can lead to the model over-fitting superior modalities while disregarding the information present in inferior modalities, making it challenging for the target model to harness the synergistic and complementary effects of multi-modality information. Therefore, we proposed a collaborative multi-step transfer method to concretize the modality's contribution, and construct the data-driven information regulation modulation to dynamically adjust the information usage of every modality at the sample level. Our method has demonstrated the superiority of our proposed method over other solutions across both simulated and real cloud-covered scene classification datasets. Additionally, we further discussed the working mechanism and limitation of our method, and analyzed the 'modality imbalance' problem. \textcolor{red}{ In the future, we aim to integrate multi-modality pre-trained models into our transfer learning framework, capitalizing on the continuous advancements in unsupervised pre-training methods to make the acquisition of prior knowledge for SAR feature recognition feasible.} Additionally, we also plan to further explore strategies to address modality imbalances under increasingly extreme conditions.



 
%

\bibliographystyle{IEEEtran} 
\bibliography{IEEEabrv,final_bib_6.12}

\begin{thebibliography}{10}
\providecommand{\url}[1]{#1}
\csname url@samestyle\endcsname
\providecommand{\newblock}{\relax}
\providecommand{\bibinfo}[2]{#2}
\providecommand{\BIBentrySTDinterwordspacing}{\spaceskip=0pt\relax}
\providecommand{\BIBentryALTinterwordstretchfactor}{4}
\providecommand{\BIBentryALTinterwordspacing}{\spaceskip=\fontdimen2\font plus
\BIBentryALTinterwordstretchfactor\fontdimen3\font minus \fontdimen4\font\relax}
\providecommand{\BIBforeignlanguage}[2]{{%
\expandafter\ifx\csname l@#1\endcsname\relax
\typeout{** WARNING: IEEEtran.bst: No hyphenation pattern has been}%
\typeout{** loaded for the language `#1'. Using the pattern for}%
\typeout{** the default language instead.}%
\else
\language=\csname l@#1\endcsname
\fi
#2}}
\providecommand{\BIBdecl}{\relax}
\BIBdecl

\bibitem{liu2023multi}
Q.~Liu, M.~He, Y.~Kuang, L.~Wu, J.~Yue, and L.~Fang, ``A multi-level label-aware semi-supervised framework for remote sensing scene classification,'' \emph{IEEE Transactions on Geoscience and Remote Sensing}, 2023.

\bibitem{ghazouani2019multi}
F.~Ghazouani, I.~R. Farah, and B.~Solaiman, ``A multi-level semantic scene interpretation strategy for change interpretation in remote sensing imagery,'' \emph{IEEE Transactions on Geoscience and Remote Sensing}, vol.~57, no.~11, pp. 8775--8795, 2019.

\bibitem{liu2017classifying}
X.~Liu, J.~He, Y.~Yao, J.~Zhang, H.~Liang, H.~Wang, and Y.~Hong, ``Classifying urban land use by integrating remote sensing and social media data,'' \emph{International Journal of Geographical Information Science}, vol.~31, no.~8, pp. 1675--1696, 2017.

\bibitem{su2023reconstruction}
Y.~Su, G.~Zhang, S.~Mei, J.~Lian, Y.~Wang, and S.~Wan, ``Reconstruction-assisted and distance-optimized adversarial training: A defense framework for remote sensing scene classification,'' \emph{IEEE Transactions on Geoscience and Remote Sensing}, 2023.

\bibitem{cheng2020remote}
G.~Cheng, X.~Xie, J.~Han, L.~Guo, and G.-S. Xia, ``Remote sensing image scene classification meets deep learning: Challenges, methods, benchmarks, and opportunities,'' \emph{IEEE Journal of Selected Topics in Applied Earth Observations and Remote Sensing}, vol.~13, pp. 3735--3756, 2020.

\bibitem{zhang2004calculation}
Y.~Zhang, W.~B. Rossow, A.~A. Lacis, V.~Oinas, and M.~I. Mishchenko, ``Calculation of radiative fluxes from the surface to top of atmosphere based on isccp and other global data sets: Refinements of the radiative transfer model and the input data,'' \emph{Journal of Geophysical Research: Atmospheres}, vol. 109, no. D19, 2004.

\bibitem{mendoza2021thermodynamics}
V.~Mendoza, M.~Pazos, R.~Gardu{\~n}o, and B.~Mendoza, ``Thermodynamics of climate change between cloud cover, atmospheric temperature and humidity,'' \emph{Scientific Reports}, vol.~11, no.~1, p. 21244, 2021.

\bibitem{yang2019object}
F.~Yang, X.~Yang, Z.~Wang, C.~Lu, Z.~Li, and Y.~Liu, ``Object-based classification of cloudy coastal areas using medium-resolution optical and sar images for vulnerability assessment of marine disaster,'' \emph{Journal of oceanology and limnology}, vol.~37, no.~6, pp. 1955--1970, 2019.

\bibitem{niu2020decade}
S.~Niu, Y.~Liu, J.~Wang, and H.~Song, ``A decade survey of transfer learning (2010--2020),'' \emph{IEEE Transactions on Artificial Intelligence}, vol.~1, no.~2, pp. 151--166, 2020.

\bibitem{wang2022empirical}
D.~Wang, J.~Zhang, B.~Du, G.-S. Xia, and D.~Tao, ``An empirical study of remote sensing pretraining,'' \emph{IEEE Transactions on Geoscience and Remote Sensing}, vol.~61, pp. 1--20, 2022.

\bibitem{cheng2017remote}
G.~Cheng, J.~Han, and X.~Lu, ``Remote sensing image scene classification: Benchmark and state of the art,'' \emph{Proceedings of the IEEE}, vol. 105, no.~10, pp. 1865--1883, 2017.

\bibitem{xia2017aid}
G.-S. Xia, J.~Hu, F.~Hu, B.~Shi, X.~Bai, Y.~Zhong, L.~Zhang, and X.~Lu, ``Aid: A benchmark data set for performance evaluation of aerial scene classification,'' \emph{IEEE Transactions on Geoscience and Remote Sensing}, vol.~55, no.~7, pp. 3965--3981, 2017.

\bibitem{ding2017learning}
Y.~Ding, Y.~Li, and W.~Yu, ``Learning from label proportions for sar image classification,'' \emph{Eurasip Journal on Advances in Signal Processing}, vol. 2017, pp. 1--12, 2017.

\bibitem{huang2020classification}
Z.~Huang, C.~O. Dumitru, Z.~Pan, B.~Lei, and M.~Datcu, ``Classification of large-scale high-resolution sar images with deep transfer learning,'' \emph{IEEE Geoscience and Remote Sensing Letters}, vol.~18, no.~1, pp. 107--111, 2020.

\bibitem{lorenzi2011inpainting}
L.~Lorenzi, F.~Melgani, and G.~Mercier, ``Inpainting strategies for reconstruction of missing data in vhr images,'' \emph{IEEE Geoscience and remote sensing letters}, vol.~8, no.~5, pp. 914--918, 2011.

\bibitem{xu2021missing}
H.~Xu, X.~Tang, B.~Ai, X.~Gao, F.~Yang, and Z.~Wen, ``Missing data reconstruction in vhr images based on progressive structure prediction and texture generation,'' \emph{ISPRS Journal of Photogrammetry and Remote Sensing}, vol. 171, pp. 266--277, 2021.

\bibitem{zhang2018missing}
Q.~Zhang, Q.~Yuan, C.~Zeng, X.~Li, and Y.~Wei, ``Missing data reconstruction in remote sensing image with a unified spatial--temporal--spectral deep convolutional neural network,'' \emph{IEEE Transactions on Geoscience and Remote Sensing}, vol.~56, no.~8, pp. 4274--4288, 2018.

\bibitem{tao2022thick}
C.~Tao, S.~Fu, J.~Qi, and H.~Li, ``Thick cloud removal in optical remote sensing images using a texture complexity guided self-paced learning method,'' \emph{IEEE Transactions on Geoscience and Remote Sensing}, vol.~60, pp. 1--12, 2022.

\bibitem{gawlikowski2022explaining}
J.~Gawlikowski, P.~Ebel, M.~Schmitt, and X.~X. Zhu, ``Explaining the effects of clouds on remote sensing scene classification,'' \emph{IEEE Journal of Selected Topics in Applied Earth Observations and Remote Sensing}, vol.~15, pp. 9976--9986, 2022.

\bibitem{shen2014effective}
H.~Shen, H.~Li, Y.~Qian, L.~Zhang, and Q.~Yuan, ``An effective thin cloud removal procedure for visible remote sensing images,'' \emph{ISPRS Journal of Photogrammetry and Remote Sensing}, vol.~96, pp. 224--235, 2014.

\bibitem{liu2018can}
L.~Liu and B.~Lei, ``Can sar images and optical images transfer with each other?'' in \emph{IGARSS 2018-2018 IEEE International Geoscience and Remote Sensing Symposium}.\hskip 1em plus 0.5em minus 0.4em\relax IEEE, 2018, pp. 7019--7022.

\bibitem{wang2022sar}
H.~Wang, Z.~Zhang, Z.~Hu, and Q.~Dong, ``Sar-to-optical image translation with hierarchical latent features,'' \emph{IEEE Transactions on Geoscience and Remote Sensing}, vol.~60, pp. 1--12, 2022.

\bibitem{song2022two}
Y.~Song, J.~Li, P.~Gao, L.~Li, T.~Tian, and J.~Tian, ``Two-stage cross-modality transfer learning method for military-civilian sar ship recognition,'' \emph{IEEE Geoscience and Remote Sensing Letters}, vol.~19, pp. 1--5, 2022.

\bibitem{senapati2023image}
R.~K. Senapati, R.~Satvika, A.~Anmandla, G.~Ashesh~Reddy, and C.~Anil~Kumar, ``Image-to-image translation using pix2pix gan and cycle gan,'' in \emph{International Conference on Data Intelligence and Cognitive Informatics}.\hskip 1em plus 0.5em minus 0.4em\relax Springer, 2023, pp. 573--586.

\bibitem{rostami2019deep}
M.~Rostami, S.~Kolouri, E.~Eaton, and K.~Kim, ``Deep transfer learning for few-shot sar image classification,'' \emph{Remote Sensing}, vol.~11, no.~11, p. 1374, 2019.

\bibitem{zhu2021deep}
Y.~Zhu, F.~Zhuang, J.~Wang, G.~Ke, J.~Chen, J.~Bian, H.~Xiong, and Q.~He, ``Deep subdomain adaptation network for image classification,'' \emph{IEEE Transactions on Neural Networks and Learning Systems}, vol.~32, no.~4, pp. 1713--1722, 2021.

\bibitem{peng2022domain}
J.~Peng, Y.~Huang, W.~Sun, N.~Chen, Y.~Ning, and Q.~Du, ``Domain adaptation in remote sensing image classification: A survey,'' \emph{IEEE Journal of Selected Topics in Applied Earth Observations and Remote Sensing}, vol.~15, pp. 9842--9859, 2022.

\bibitem{attarchi2020extracting}
S.~Attarchi, ``Extracting impervious surfaces from full polarimetric sar images in different urban areas,'' \emph{International Journal of Remote Sensing}, vol.~41, no.~12, pp. 4644--4663, 2020.

\bibitem{dos2021vegetation}
E.~P. dos Santos, D.~D. Da~Silva, and C.~H. do~Amaral, ``Vegetation cover monitoring in tropical regions using sar-c dual-polarization index: seasonal and spatial influences,'' \emph{International Journal of Remote Sensing}, vol.~42, no.~19, pp. 7581--7609, 2021.

\bibitem{amarsaikhan2010fusing}
D.~Amarsaikhan, H.~Blotevogel, J.~Van~Genderen, M.~Ganzorig, R.~Gantuya, and B.~Nergui, ``Fusing high-resolution sar and optical imagery for improved urban land cover study and classification,'' \emph{International Journal of Image and Data Fusion}, vol.~1, no.~1, pp. 83--97, 2010.

\bibitem{bai2021comprehensively}
Y.~Bai, G.~Sun, Y.~Li, P.~Ma, G.~Li, and Y.~Zhang, ``Comprehensively analyzing optical and polarimetric sar features for land-use/land-cover classification and urban vegetation extraction in highly-dense urban area,'' \emph{International Journal of Applied Earth Observation and Geoinformation}, vol. 103, p. 102496, 2021.

\bibitem{mahyoub2019fusing}
S.~Mahyoub, A.~Fadil, E.~Mansour, H.~Rhinane, and F.~Al-Nahmi, ``Fusing of optical and synthetic aperture radar (sar) remote sensing data: A systematic literature review (slr),'' \emph{The International Archives of the Photogrammetry, Remote Sensing and Spatial Information Sciences}, vol.~42, pp. 127--138, 2019.

\bibitem{wang2021knowledge}
L.~Wang and K.-J. Yoon, ``Knowledge distillation and student-teacher learning for visual intelligence: A review and new outlooks,'' \emph{IEEE transactions on pattern analysis and machine intelligence}, vol.~44, no.~6, pp. 3048--3068, 2021.

\bibitem{wang2021cross}
Y.~Wang, R.~Xiao, J.~Qi, and C.~Tao, ``Cross-sensor remote-sensing images scene understanding based on transfer learning between heterogeneous networks,'' \emph{IEEE Geoscience and Remote Sensing Letters}, vol.~19, pp. 1--5, 2021.

\bibitem{tao2023general}
C.~Tao, R.~Xiao, Y.~Wang, J.~Qi, and H.~Li, ``A general transitive transfer learning framework for cross-optical sensor remote sensing image scene understanding,'' \emph{IEEE Journal of Selected Topics in Applied Earth Observations and Remote Sensing}, vol.~16, pp. 4248--4260, 2023.

\bibitem{gou2021knowledge}
J.~Gou, B.~Yu, S.~J. Maybank, and D.~Tao, ``Knowledge distillation: A survey,'' \emph{International Journal of Computer Vision}, vol. 129, no.~6, pp. 1789--1819, 2021.

\bibitem{geirhos2020shortcut}
R.~Geirhos, J.-H. Jacobsen, C.~Michaelis, R.~Zemel, W.~Brendel, M.~Bethge, and F.~A. Wichmann, ``Shortcut learning in deep neural networks,'' \emph{Nature Machine Intelligence}, vol.~2, no.~11, pp. 665--673, 2020.

\bibitem{wang2020makes}
W.~Wang, D.~Tran, and M.~Feiszli, ``What makes training multi-modal classification networks hard?'' in \emph{Proceedings of the IEEE/CVF conference on computer vision and pattern recognition}, 2020, pp. 12\,695--12\,705.

\bibitem{peng2022balanced}
X.~Peng, Y.~Wei, A.~Deng, D.~Wang, and D.~Hu, ``Balanced multimodal learning via on-the-fly gradient modulation,'' in \emph{Proceedings of the IEEE/CVF conference on computer vision and pattern recognition}, 2022, pp. 8238--8247.

\bibitem{chauhan2016comparative}
S.~Chauhan and H.~S. Srivastava, ``Comparative evaluation of the sensitivity of multi-polarized sar and optical data for various land cover classes,'' \emph{Int. J. Adv. Remote Sens. GIS Geogr}, vol.~4, no.~1, pp. 1--14, 2016.

\bibitem{schmitt2019sen12ms}
M.~Schmitt, L.~H. Hughes, C.~Qiu, and X.~X. Zhu, ``Sen12ms--a curated dataset of georeferenced multi-spectral sentinel-1/2 imagery for deep learning and data fusion,'' \emph{arXiv preprint arXiv:1906.07789}, 2019.

\bibitem{zhang2018mapping}
X.~Zhang, B.~Wu, G.~E. Ponce-Campos, M.~Zhang, S.~Chang, and F.~Tian, ``Mapping up-to-date paddy rice extent at 10 m resolution in china through the integration of optical and synthetic aperture radar images,'' \emph{Remote Sensing}, vol.~10, no.~8, p. 1200, 2018.

\bibitem{he2016deep}
K.~He, X.~Zhang, S.~Ren, and J.~Sun, ``Deep residual learning for image recognition,'' in \emph{Proceedings of the IEEE conference on computer vision and pattern recognition}, 2016, pp. 770--778.

\bibitem{szegedy2016rethinking}
C.~Szegedy, V.~Vanhoucke, S.~Ioffe, J.~Shlens, and Z.~Wojna, ``Rethinking the inception architecture for computer vision,'' in \emph{Proceedings of the IEEE conference on computer vision and pattern recognition}, 2016, pp. 2818--2826.

\bibitem{li2022deep}
J.~Li, D.~Hong, L.~Gao, J.~Yao, K.~Zheng, B.~Zhang, and J.~Chanussot, ``Deep learning in multimodal remote sensing data fusion: A comprehensive review,'' \emph{International Journal of Applied Earth Observation and Geoinformation}, vol. 112, p. 102926, 2022.

\bibitem{huang2021makes}
Y.~Huang, C.~Du, Z.~Xue, X.~Chen, H.~Zhao, and L.~Huang, ``What makes multi-modal learning better than single (provably),'' \emph{Advances in Neural Information Processing Systems}, vol.~34, pp. 10\,944--10\,956, 2021.

\bibitem{fan2023pmr}
Y.~Fan, W.~Xu, H.~Wang, J.~Wang, and S.~Guo, ``Pmr: Prototypical modal rebalance for multimodal learning,'' in \emph{Proceedings of the IEEE/CVF Conference on Computer Vision and Pattern Recognition}, 2023, pp. 20\,029--20\,038.

\bibitem{wang2023imbalance}
Y.~Wang, Y.~Wan, Y.~Zhang, B.~Zhang, and Z.~Gao, ``Imbalance knowledge-driven multi-modal network for land-cover semantic segmentation using aerial images and lidar point clouds,'' \emph{ISPRS Journal of Photogrammetry and Remote Sensing}, vol. 202, pp. 385--404, 2023.

\bibitem{yang2019deep}
Y.~Yang, Y.-F. Wu, D.-C. Zhan, Z.-B. Liu, and Y.~Jiang, ``Deep robust unsupervised multi-modal network,'' in \emph{Proceedings of the AAAI Conference on Artificial Intelligence}, vol.~33, no.~01, 2019, pp. 5652--5659.

\bibitem{liu2021contrastive}
Y.~Liu, Q.~Fan, S.~Zhang, H.~Dong, T.~Funkhouser, and L.~Yi, ``Contrastive multimodal fusion with tupleinfonce,'' in \emph{Proceedings of the IEEE/CVF International Conference on Computer Vision}, 2021, pp. 754--763.

\bibitem{sun2021learning}
Y.~Sun, S.~Mai, and H.~Hu, ``Learning to balance the learning rates between various modalities via adaptive tracking factor,'' \emph{IEEE Signal Processing Letters}, vol.~28, pp. 1650--1654, 2021.

\bibitem{li2023boosting}
H.~Li, X.~Li, P.~Hu, Y.~Lei, C.~Li, and Y.~Zhou, ``Boosting multi-modal model performance with adaptive gradient modulation,'' in \emph{Proceedings of the IEEE/CVF International Conference on Computer Vision}, 2023, pp. 22\,214--22\,224.

\bibitem{jiang2023understanding}
Q.~Jiang, C.~Chen, H.~Zhao, L.~Chen, Q.~Ping, S.~D. Tran, Y.~Xu, B.~Zeng, and T.~Chilimbi, ``Understanding and constructing latent modality structures in multi-modal representation learning,'' in \emph{Proceedings of the IEEE/CVF Conference on Computer Vision and Pattern Recognition}, 2023, pp. 7661--7671.

\bibitem{bony2015clouds}
S.~Bony, B.~Stevens, D.~M. Frierson, C.~Jakob, M.~Kageyama, R.~Pincus, T.~G. Shepherd, S.~C. Sherwood, A.~P. Siebesma, A.~H. Sobel \emph{et~al.}, ``Clouds, circulation and climate sensitivity,'' \emph{Nature Geoscience}, vol.~8, no.~4, pp. 261--268, 2015.

\bibitem{dou2020unpaired}
Q.~Dou, Q.~Liu, P.~A. Heng, and B.~Glocker, ``Unpaired multi-modal segmentation via knowledge distillation,'' \emph{IEEE transactions on medical imaging}, vol.~39, no.~7, pp. 2415--2425, 2020.

\bibitem{qian2021hybrid}
X.~Qian, F.~Liu, L.~Jiao, X.~Zhang, P.~Chen, L.~Li, J.~Gu, and Y.~Cui, ``A hybrid network with structural constraints for sar image scene classification,'' \emph{IEEE Transactions on Geoscience and Remote Sensing}, vol.~60, pp. 1--17, 2021.

\end{thebibliography}


\section{Biography Section}

\vspace{-33pt}
\begin{IEEEbiography}
[{\includegraphics[width=1in,height=1.25in,clip,keepaspectratio]{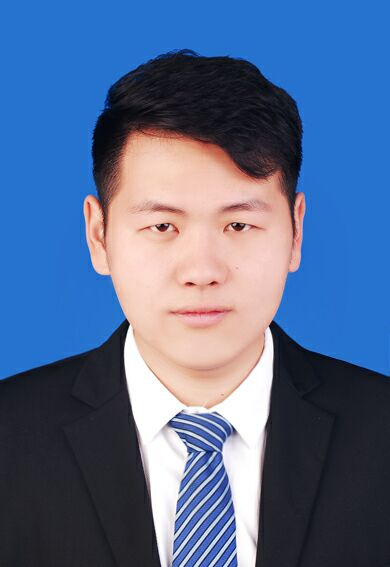 }}]{Yuze Wang} 
received the B.E. degree in surveying and mapping in 2020 from the School of Geosciences and Info-Physics, Central South University, Changsha, China, where he is currently working toward a Ph.D. degree in surveying and mapping. His main inquiry direction is to solve the existing problems of deep learning techniques in the fields of semantic segmentation, scene classification, and edge detection. His research interests include the application of deep learning techniques and remote sensing data in the field of agriculture.
\end{IEEEbiography}

\begin{IEEEbiography}
[{\includegraphics[width=1in,height=1.25in,clip,keepaspectratio]{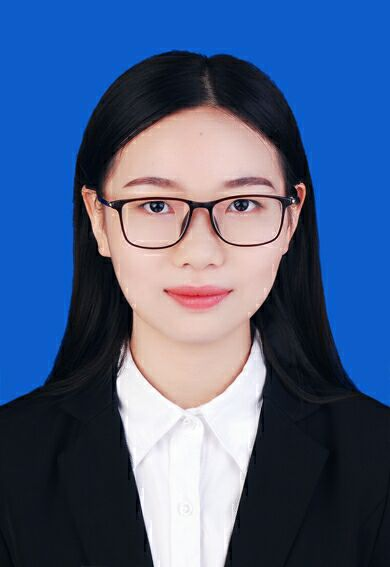 }}]{Rong Xiao} 
has completed her M.S. degree in surveying and mapping at the School of Geosciences and Info-Physics, Central South University, Changsha, China. Her research interests encompass remote sensing, computer vision, deep learning, transfer learning, and knowledge distillation
\end{IEEEbiography}

\begin{IEEEbiography}
[{\includegraphics[width=1in,height=1.25in,clip,keepaspectratio]{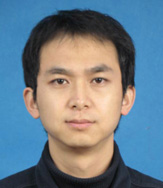 }}]{Haifeng Li} 
 (Member, IEEE) received a master’s degree in transportation engineering from the South China University of Technology, Guangzhou, China, in 2005, and a Ph.D. degree in photogrammetry and remote sensing from Wuhan University, Wuhan, China, in 2009. He is currently a Profesatwith the School of Geosciences and Info-Physics, Central South University, Changsha, China. He was a Research Associate with the Department of Land Surveying and GeoInformatics, The Hong Kong Polytechnic University, Hong Kong, in 2011, and a Visiting Scholar with the University of Illinois at Urbana-Champaign, Urbana, IL, USA, from 2013 to 2014. He has authored over 30 journal papers. His current research interests include geo/remote sensing Big Data, machine/deep learning, and artificial/brain-inspired intelligence. Dr. Li is a reviewer for many journals.
\end{IEEEbiography}

\begin{IEEEbiography}
[{\includegraphics[width=1in,height=1.25in,clip,keepaspectratio]{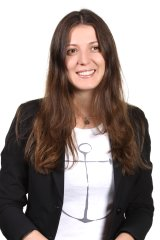 }}]{Mariana Belgiu} 
 received aMSc degree in Geoinformatics and a Ph.D. degree in remote sensing all from the University of Salzburg. She is currently an Associate Professor in remote sensing at the Faculty of Geo-Information Science and Earth Observation (ITC), University of Twente. Her research is mainly focused on developing and applying cutting-edge machine learning and deep learning methods to analyze multi-temporal multispectral and hyperspectral remote sensing imagery for generating spatially explicit information to address pressing environmental problems, including agriculture mapping and monitoring and assessing the nutritional quality of crops.  
\end{IEEEbiography}

\begin{IEEEbiography}
[{\includegraphics[width=1in,height=1.25in,clip,keepaspectratio]{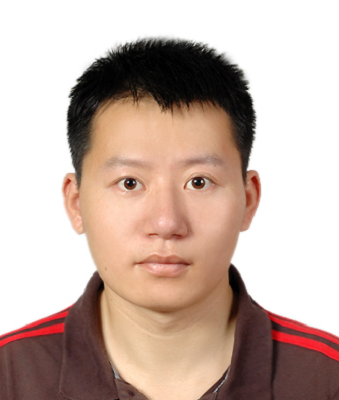}}]{Chao Tao} 
received a B.S. degree in computational mathematics from the School of Mathematics and Statistics, Huazhong University of Science and Technology, Wuhan, China, in 2007, and the Ph.D. degree in pattern recognition and intelligent system from the Institution of Pattern Recognition and Artificial Intelligence, Huazhong University of Science and Technology, Wuhan, China, in 2012. He is currently a Profesatwith the School of Geosciences and Info-Physics, Central South University, Changsha, China. He has authored more than 30 peer-reviewed articles in international journals from multiple domains such as remote sensing and computer vision. His research interests include computer vision, machine learning, deep learning, and their applications in remote sensing. Dr. Tao has frequently served as a reviewer for several international journals including the IEEE-TGRS, IEEE-GRSL, IEEE-JSTAR, PERS, and ISPRS JPRS. He is also a Communication Evaluation Expert for the National Natural Science Foundation of China.
\end{IEEEbiography}

\vspace{-33pt}

\vfill

\end{document}